\documentclass[sn-basic,iicol]{sn-jnl}

\jyear{2021}%

\begin{document}

\title[Revisiting Point Cloud Simplification]{\centering Revisiting Point Cloud Simplification: \\ 
A  Learnable Feature Preserving Approach}


\author*[1]{\fnm{Rolandos-Alexandros} \sur{Potamias}}\email{r.potamias@imperial.ac.uk}

\author[1]{\fnm{Giorgos} \sur{Bouritsas}}\email{g.bouritsas@imperial.ac.uk}

\author[1]{\fnm{Stefanos} \sur{Zafeiriou}}\email{s.zafeiriou@imperial.ac.uk}

\affil*[1]{\orgdiv{Department of Computing}, \orgname{Imperial College London}, \orgaddress{ \city{London},  \country{UK}}}


\abstract{The recent advances in 3D sensing technology have made possible the capture of point clouds in significantly high resolution. However, increased detail usually comes at the expense of high storage, as well as computational costs in terms of processing and visualization operations. Mesh and Point Cloud simplification methods aim to reduce the complexity of 3D models while retaining visual quality and relevant salient features. Traditional simplification techniques usually rely on solving a time-consuming optimization problem, hence they are impractical for large-scale datasets. In an attempt to alleviate this computational burden, we propose a fast point cloud simplification method by learning to sample salient points. The proposed method relies on a graph neural network architecture trained to select an arbitrary, user-defined, number of points from the input space and to re-arrange their positions so as to minimize the visual perception error.  The approach is extensively evaluated on various datasets using several perceptual metrics. Importantly,  our method is able to generalize to out-of-distribution shapes,  hence demonstrating zero-shot capabilities.}





\maketitle

\section{Introduction}
The progress in sensing technologies has significantly expedited the 3D data acquisition pipelines which in turn has increased the availability of large and diverse 3D datasets. With a single 3D sensing device \citep{lu2006single}, one can capture a target surface and represent it as a 3D object, with point clouds and meshes being the most popular representations. Several applications, ranging from virtual reality and 3D avatar generation \cite{lattas2020avatarme,potamias2020learning} to 3D printing and digitization of cultural heritage \citep{pavlidis2007methods}, require such representations. However, generally, a 3D capturing device generates thousands of points per second, making processing, visualization and storage of captured 3D objects a computationally daunting task. Often, raw point sets contain an enormous amount of, not only redundant, but also possibly noisy, points with low visual perceptual importance, which results into an unnecessary increase in the storage costs. Thus processing, rendering and editing applications require the development of efficient simplification methods that discard excessive details and reduce the size of the object, while preserving their significant visual characteristics.

Point cloud simplification can be described as a process of reducing the levels-of-detail (LOD) so as to minimise the introduced perceptual error \citep{cignoni1998comparison}.  In contrast to sampling methods, the main objective of point cloud simplification, is the removal or the collapse of particular points that does not significantly affect the visual perceptual quality, in a way that the most salient features are preserved in the simplified point cloud \citep{luebke2001developer}. In this study, we propose a learnable strategy to remove the least perceptually important points without sacrificing the overall structure of the point cloud. As perceptually important features that should be preserved, we consider points with increased surface curvature, that have been shown to highly correlate with the human perceptual system \citep{lee2005mesh,lavoue2009local}.   

Traditional simplification methods address the task by attempting to solve an optimization problem that minimizes the visual error of the simplified model. Usually, such optimizations are non-convex with high computational cost, where a point importance queue is constructed to sort the 3D points according to their scores \citep{rossignac1993multi,hoppe1996progressive,garland1997surface}. Point cloud simplification methods can be categorized as \textit{mesh-based} and as \textit{point decimation-based}. Mesh-based methods attempt to reconstruct a 3D surface from the point cloud and simplify the generated mesh. In contrast, point decimation-based methods directly select points from the reference point cloud according to their feature scores. Edge contraction remains to date one of the most successful and popular mesh-based methods, since it produces high quality approximations of the input \citep{garland1997surface,garland1998simplifying}. 
Although several approaches proposed parallel GPU computations to reduce the execution time even by 20 times \citep{decoro2007real,wang2019fast}, the simplification task can be still considered as a computationally hard problem. To this end, it is essential to reduce the computation and time complexity by leveraging neural networks to efficiently simplify point clouds. 

In this study, 
we propose the first, to the best of our knowledge, learnable point cloud simplification method. The proposed method preserves both the salient features as well as the overall structure of the input and can be used for real-time point cloud simplification without any prior surface reconstruction. 
We also show the limitations of popular distance metrics, such as Chamfer and Haussdorf, to capture salient details of the simplified models and we propose several evaluation criteria that are well-suited for simplification tasks. The proposed method is extensively evaluated in a series of wide range experiments.

The rest of the paper is structured as follows. In Section 2, we succinctly present a summary of related work covering the relevant areas of mesh and point cloud simplification,  point cloud sampling as well as learnable graph pooling methods. In Section 3, we present the preliminaries and the details of the proposed methods including the model architecture components, the training procedure, the limitations of uniform distance measures along with the implementation details. Section 4 is dedicated to review and present the evaluation criteria used to measure the performance of the proposed method. Finally, in Section 5, we extensively evaluate our method with a series of qualitative and quantitative experiments. In particular, we report the performance and the execution time of the proposed method using many perceptual and distance measures. In addition, we show that the simplified point clouds can be still identified by pre-trained classifiers. We also qualitatively evaluate the proposed method under noisy and in-the-wild point clouds and establish our findings using a user-study. 

\section{Related Work}

\subsection{Mesh Simplification}
\label{sec:related}
Mesh simplification is a well studied field with long history of research. A mesh can be simplified progressively by two techniques, namely vertex decimation and edge collapse. Although the first one is more interpretable, it requires re-tessellation in order to fill the generated holes. In the general case, each vertex is assigned with an importance score, which may indicate its distance from the average plane \citep{schroeder1992decimation}, its perceptual importance \citep{rossignac1993multi}, or high curvature \citep{li2014improved}, ensuring that vertices at smooth regions will be decimated before vertices with sharp features. Edge folding was first introduced in the seminal work of \citet{hoppe1996progressive}, where several transforms such as edge swap, edge split and edge collapse were introduced to minimize a simplification energy function. \citet{ronfard1996full} associated each vertex with the set of planes in its 1-hop neighborhood and defined the edge collapse cost as the maximum distance between the resulting vertex and the collapsed points' planes. \citet{garland1997surface} observed that each plane can be expressed by a fundamental \emph{quadric} matrix and they utilized an additive rule to express the distance of a point from the respective set of planes, i.e., the distance of a point from a set of planes can be expressed using the sum of their quadrics. Based on this observation, the authors proposed a Quadric Error Metric (QEM) to measure the error introduced when an edge collapses and simplified the edges with the minimum error. In a follow up work \citep{garland1998simplifying}, the authors generalized the quadric matrix to handle high dimensional attributes. \citet{cohen2004variational} observed that greedy simplification methods lead to sub-optimal meshes and attempted to tackle mesh simplification as a global optimization problem using shape proxies. In particular, they introduced a normal deviation error metric to partition the input mesh to non-overlapping connected regions and then fit plane approximations (shape proxies) to each partition. Although the process produces more accurate shape approximations of the input, the method is not particularly efficient. 

QEM simplification remains to date one of the most common techniques of mesh simplification, with several modification of the cost function to incorporate curvature features \citep{kim1999lod,kim2002surface,liu2005edge,yao2015quadratic} and preserve boundary constrains \citep{Bahirat}. Several approaches have been proposed for spectral mesh coarsening, by collapsing edges that constrain the Laplacian of the simplified mesh to be close with the original \citep{LiuSpectral,LescoatSpectral}. Recently, \citet{hanocka2019meshcnn} proposed a learnable edge collapse method that learns the importance of each edge, for task-driven mesh simplification. However, the edges are contracted in an inefficient iterative way and the resulting mesh faces can only be decimated approximately by two. The point cloud simplification methodology presented in this paper attempts to address and overcome the inefficiencies of the aforementioned approaches using a learnable alternative that works with arbitrary, user-defined decimation factors.

\subsection{Point Cloud Simplification and Sampling}
Similar to mesh simplification, iterative point selection and clustering techniques have also been proposed for point clouds \citep{pauly2002efficient,shi2011adaptive,han2015point,zhang2019feature}. In particular, \textit{point cloud simplification} can be addressed either via mesh simplification where the points are fitted to a surface and then simplified using traditional mesh simplification objectives \citep{alexa2001point,galantucci2005multilevel}, or via direct optimization on the point cloud where the points are selected and decimated according to their estimated local properties \citep{pauly2002efficient,shi2011adaptive,leal2017linear,zhang2019feature}. However, similar to the aforementioned approaches, computationally expensive iterative optimization is needed, which therefore makes them inefficient for large scale point clouds. 

In a different line of research, \textit{sampling methods} rely on a point selection scheme that focuses on retaining the overall structure of the object, instead of its salient features. Among them, Farthest Point Sampling (FPS) \citep{eldar1997farthest} remains the most popular choice and has been widely used as a building block in deep learning pipelines \citep{qi2017pointnet,pointnet++}. 
Recently, S-Net \citep{dovrat2019learning} was proposed as a learnable alternative for task-driven sampling, optimized for downstream tasks. To ensure that the selected points lie on the surface of the input point cloud, a matching step is performed using k-nearest neighbors. In a follow-up work, SampleNet \citep{lang2020samplenet} extended S-Net by introducing a differentiable relaxation to the nearest neighbour matching step. Although the learnable sampling studies are closely related to ours, they only sample point clouds in a task-driven manner and as a result the preservation of the high frequency details of the point cloud is not ensured.
\begin{figure*}[!ht]
\begin{center}

   \includegraphics[width=0.75\linewidth]{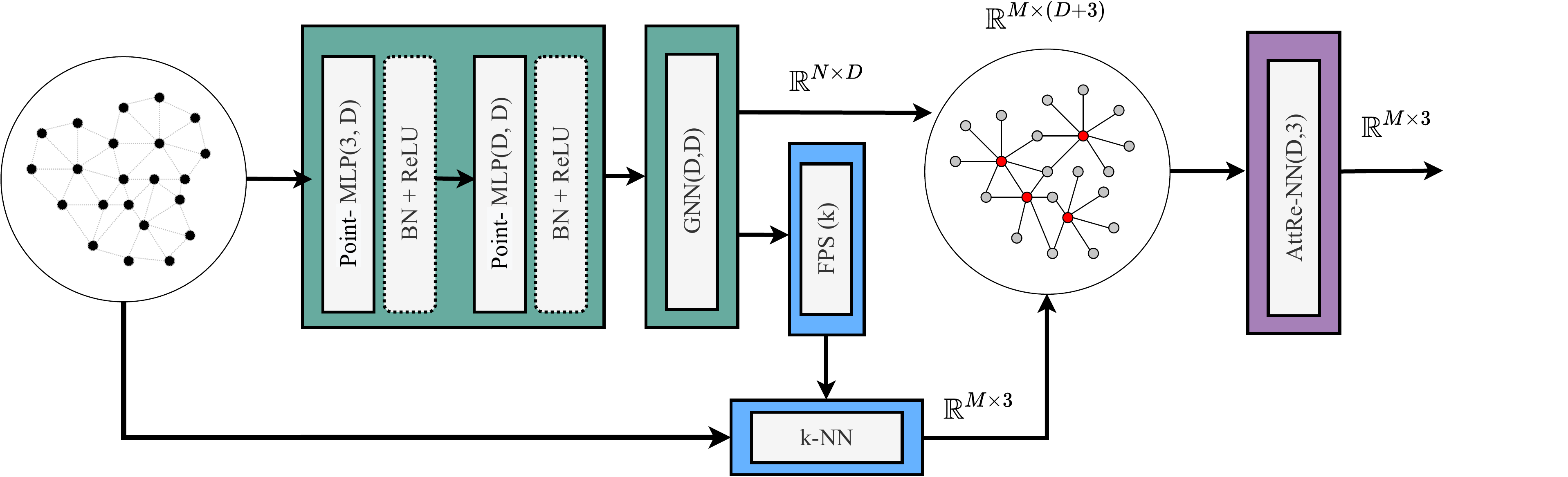}
\end{center}
   \caption{Overview of the proposed method. Initially a point cloud with optional triangulation (mesh) is passed through a projection network (green) and embedded to a higher dimensional latent space. FPS is used to select points from the set of latent representations (blue) that can be conceived as cluster centers of the input. Finally, a k-NN graph is constructed between the cluster centers and the input points that is used to modify their positions using the refinement layer (purple).}
\label{fig:model}
\end{figure*}

\subsection{Graph Pooling}
One of the components underlying the success of large-scale CNNs are pooling layers, introduced to formulate training in a hierarchical manner. Until recently, graph neural network (GNN) architectures used for tasks like classification, segmentation and generation, learn global graph representations by relying solely on node aggregations, neglecting the characteristics of local substructures. To mitigate such issues, several graph pooling layers have been introduced for hierarchical representation learning. Initial approaches, utilized variations of the Graclus clustering algorithm \citep{dhillon2007weighted,DefferrardNIPS2016,fey2018splinecnn} to perform pooling operations on the input node set. The first differentiable pooling layer (DiffPool) was introduced by \citet{YingNEURIPS2018} that learns a soft assignment matrix to perform node clustering. However, the clustering assignment matrix requires quadratic storage complexity and it is not scalable to large scale graphs \citep{cangea2018towards}. To address the limitations of DiffPool several Top-K selection methods have been proposed, that select the top ranking nodes according to a learnable projection score \citep{gao2019graph,cangea2018towards}. In order to enrich the projection score with local graph structure, SAGPool \citep{pmlr-v97-lee19c} utilized a GNN layer to assign self-attention scores to each node. Recently, \citet{ranjan2019asap} introduced ASAPooling, an extension to Top-k pooling schemes that performs node aggregation to address the edge connectivity limitations of the previous methods. However, Top-k selection methods cannot generalize under different selections of \textit{k}, which is important for simplification tasks, and therefore their applicability is limited to graph pooling layers for hierarchical learning.

\subsection{Assessment of Perceptual Visual Quality}
Processes such as simplification, lossy compression, watermarking and filtering inevitably introduce distortion to the 3D objects. Minimizing and measuring the visual cost in rendered data is a long studied problem \citep{lavoue2010comparison,corsini2013perceptual}. Inspired by Image Quality Assessment measures, the objective of Perceptual Visual Quality (PVQ) assessment is to measure the distortion introduced compared to the original version of the object. The most important factor of the PVQ assessment is to correlate with the Human Perceptual System (HPS). Initially, 3D PVQ used to be estimated using two-dimensional IQA measures of the rendered models \citep{lindstrom2000out,qu2008perceptually}. However, these methods neglect the depth perceptual quality, which is an important factor to 3D PVQ. To address this limitation, several methods have been proposed acting directly on 3D positions and measuring the PVQ as a function of Laplacian distances \citep{karni2000spectral}, curvature statistics \citep{kim2002discrete,lavoue2011multiscale, torkhani2012curvature}, dihedral angles \citep{vavsa2012dihedral} or per vertex roughness \cite{corsini2013perceptual}. \cite{lavoue2013perceptual} attempted to identify the most relevant geometric attributes for mesh PVQ using scores derived from subjective assessments. In particular, the authors selected a variety of perceptual attributes presented in the literature and utilized two datasets with users' opinion scores to learn the attributes that correlate with human perception. The findings showed that curvature related features such as min/max/mean curvature along with dihedral angles relate the most with human perception. Several follow-up studies have shown that curvature features reflect human perception \citep{dong2015perceptual,feng2018perceptual}. Recently, \cite{yildiz2020machine} also explored the importance of geometric attributes in PVQ, using ground truth labels gathered from a crowdsourcing platform. Their results coincided with the finding of previous studies, indicating that curvature and roughness are strong indicators of similarity in the HVS.
In this work, motivated by the aforementioned studies we utilized curvature related losses and quality measures to train and assess the performance of the proposed model. Throughout this paper we will refer to perceptual metrics as the ones that relate with 3D curvature. 

\section{Method}
\label{sec:method}
\subsection{Preliminaries: Point Curvature Estimation}
\label{sec:normalest}
Calculating the local surface properties of an unstructured point cloud is a non-trivial problem. As demonstrated in  \citep{hoppe1992surface,pauly2002efficient},  covariance analysis can be an intuitive estimator of the surface normals and curvature. In particular, considering a neighborhood $\mathcal{N}_i$ around the point $\mathbf{p}_i \in \mathcal{R}^3$ we can define the covariance matrix: 

\begin{equation}
  C =\begin{bmatrix}
    \mathbf{p}_{i_1} - \mathbf{p}_i \\
    \mathbf{p}_{i_2} - \mathbf{p}_i\\
    \vdots \\
    \mathbf{p}_{i_k} - \mathbf{p}_i
\end{bmatrix}^T \cdot \begin{bmatrix}
    \mathbf{p}_{i_1} - \mathbf{p}_i \\
    \mathbf{p}_{i_2} - \mathbf{p}_i\\
    \vdots \\
    \mathbf{p}_{i_k} - \mathbf{p}_i
\end{bmatrix} 
\in \mathbb{R}^{\lvert\mathcal{N}_i\rvert\times\lvert\mathcal{N}_i\lvert}  
\end{equation}
where $\mathbf{p}_{i_j} \in \mathcal{N}_i$.

Solving the eigendecomposition of the covariance matrix $C$, we can derive the eigenvectors corresponding to the principal eigenvalues, that define an orthogonal frame at point $\mathbf{p}_i$. The eigenvalues $\lambda_i$ measure the variation along the axis defined by their corresponding eigenvector. Intuitively, the eigenvectors that correspond to the largest eigenvalues span the tangent plane at point $\mathbf{p}_i$, whereas the eigenvector corresponding to the smallest eigenvalue can be used to approximate the surface normal $n_i$. Thus, given that the smallest eigenvalue measures the deviation of point $\mathbf{p}_i$ from the surface, it can be used as an estimate of point curvature. As shown in \citep{pauly2002efficient}, we may define:
\begin{equation}
   \kappa(\mathbf{p}_i) = \frac{\lambda_0}{\lambda_0+\lambda_1+\lambda_2},    \quad     \lambda_0<\lambda_1<\lambda_2 
\end{equation}
as the local curvature estimate at point $\mathbf{p}_i$ which is ideal for tasks such as point simplification. Using the previously estimated curvature at point $\mathbf{p}_i$ we can estimate the mean curvature as the Gaussian weighted average of the curvatures around the neighborhood $\mathcal{N}_i$: 
\begin{equation}
    \bar{\mathcal{K}}(\mathbf{p}_i) = \frac{\sum\limits_{j \in \mathcal{N}_i} \kappa(\mathbf{p}_j) \exp{(-\|\mathbf{p}_j - \mathbf{p}_i\|^2/h})}{\sum\limits_{j \in \mathcal{N}_i}\exp{(-\|\mathbf{p}_j - \mathbf{p}_i\|^2/h}) } 
\end{equation}
where $h$ is a constant defining the radius of the neighborhood. Finally, we can define an estimation of the roughness as the difference between curvature and the mean curvature at point $\mathbf{p}_i$ as: 
\begin{equation}
    \mathcal{R}(\mathbf{p}_i)=\lvert\kappa(\mathbf{p}_i) -\bar{\mathcal{K}}(\mathbf{p}_i)\rvert
\end{equation}

\subsection{Model}
\begin{figure}[t]
\begin{center}

   \includegraphics[width=0.6\linewidth]{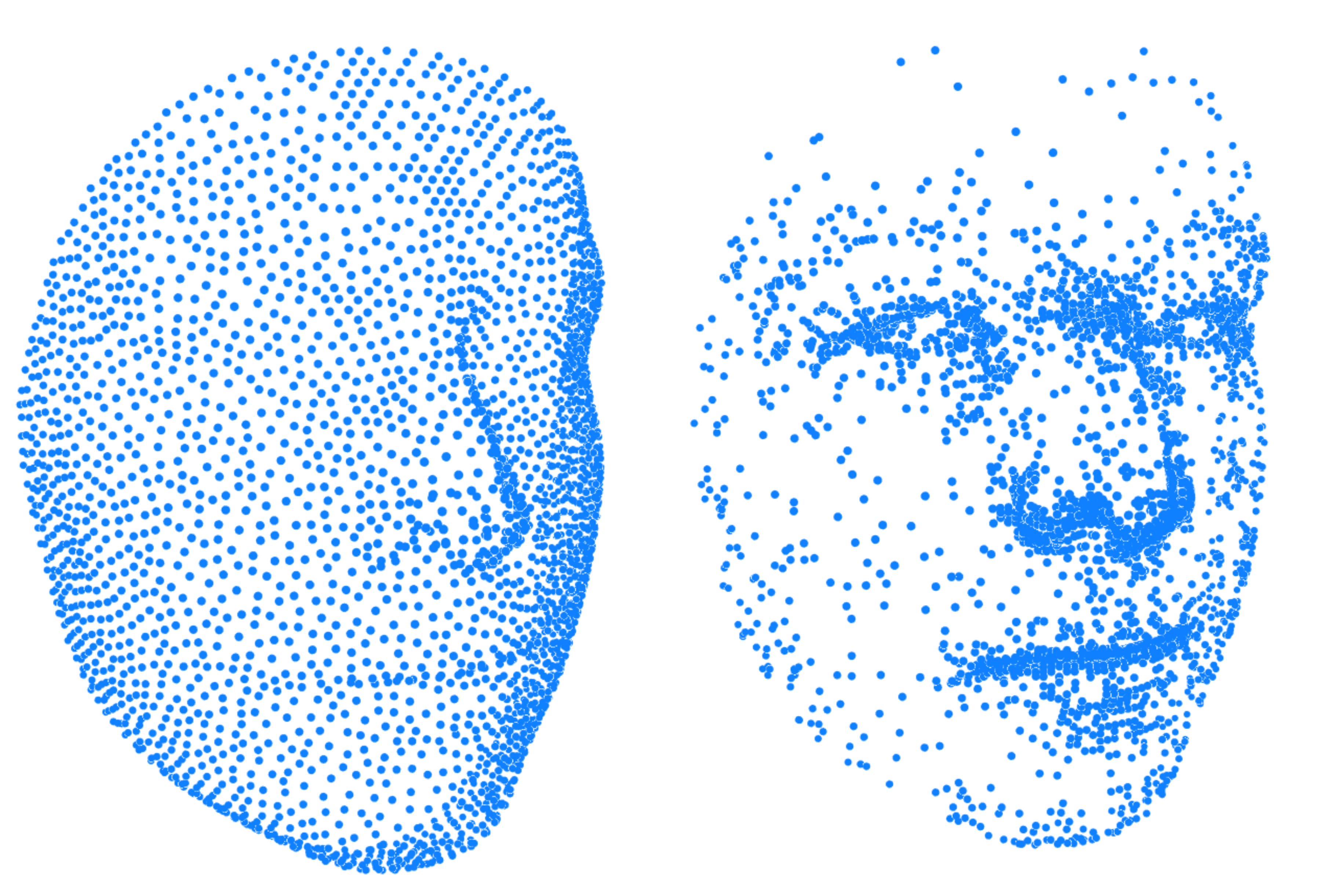}
\end{center}
   \caption{Point cloud simplified using FPS (left) achieves better Chamfer distance (CD) than a point cloud decimated using curvature-preservation methods (right). However, the perceptual similarity scores are better for the latter.}
\label{fig:smoothcd}
\end{figure}
The main building block of our architecture is a graph neural network that receives at its input a point cloud (or a mesh) $\mathcal{P}_1$ with $N$ points $\mathbf{p}_i$ and outputs a simplified version $\mathcal{P}_2$ with $M$ points, $M<<N$. It is important to note that the simplified point cloud $\mathcal{P}_2$  do not need to be a subset of the original point set $\mathcal{P}_1$.  The proposed model is composed by three modules: the \textit{Projection Network}, the \textit{Point Selector} and the \textit{Refinement Network}. Figure \ref{fig:model} illustrates the architecture of the proposed method.
\subsubsection{Projection Network and Point Selector}
Point cloud simplification can be considered as a sampling procedure constrained to preserve both the overall shape and the salient features of the input cloud. In this study, we attempted to formulate sampling as a clustering problem. In particular, we aim to cluster points that share similar perceptual and structural features and express the simplified point cloud using the cluster centres. To do so, we designed a \textit{Projector Network} that embeds $(x,y,z)$ coordinates to a high dimensional space, where points with similar features will be close in the latent space. In other words, instead of directly sampling from the Euclidean input space, we aim to sample points embedded to a latent space that captures the perceptual characteristics of the input cloud. Clustering the latent space will create clusters with latent vectors of points that share similar perceptual characteristics. 

Based on the observations that Farthest Point Sampling (FPS) provides a simple and intuitive technique to select points covering the point cloud structure \citep{pointnet++}, we built a sampling module on top of this sampling strategy, where points are sampled from a high dimensional space instead of the input \textit{xyz}-space. Although any clustering algorithm could be adequate, we utilized FPS module since it covers sufficiently the input space without solving any optimization problem. Intuitively, using this formulation we are allowed to interfere the selection process and transform it to a learnable module.  The revised sampling module will select point embeddings that cover the perceptual latent space, enabling the preservation of both the shape and the features of the input. 

Projector Network comprises of a multi-layer perceptron (MLP) applied to each point independently, followed by a GNN that captures the local geometric properties around each point. The update rule of the GNN layer is the following:
\begin{equation}
\mathbf{f}_i' =  W_c \mathbf{f}_i +\frac{1}{\mathcal{N}_i} \sum_{j \in\mathcal{N}_i }  W_n \mathbf{f}_j
\end{equation}
where $\mathbf{f}_i$ denotes the output of the shared point-wise MLP for point  $\mathbf{p}_i$ and $W_c, W_n$ represent learnable projection matrices. The connectivity between points can be given either by the mesh triangulation or by a k-nn query in the input space. Following the Projector Network, \textit{Point Selector} module utilizes FPS to select points, i.e. cluster centers, based on their latent features, in order to cover the latent space.  
Given the cluster centers selected by FPS, we design a  nearest neighbour graph that connects the center points of the input with their k-nearest neighbours.  
In order to gain flexibility in cluster center positioning and preserve salient features we have selected a large enough neighborhood size.

\subsubsection{Attention-based Refinement Layer}
 Cluster centers, their neighboring point positions  along with their respective embeddings from the projection networks are fed to the attention-based refinement layer (AttRef) that modifies the positions of the cluster centers.  This layer can be considered as a rectification step that given a large neighborhood and its corresponding latent features, displaces the cluster center points in order to minimize the visual perceptual error. Given that the latent embeddings of each point can be thought of as a local descriptor of the point, the refinement layer generates the new positions based on the vertex displacements along with the neighborhood local descriptors. The final positions of the points as predicted by \textit{AttRef} are defined as follows: 

\begin{equation}
\mathbf{p}_{c_i}' =  \mathbf{p}_{c_i} +  \gamma \left( \frac{1}{\mathcal{N}_{c_i}}\sum_{j \in
\mathcal{N}_{c_i}} \alpha_{ij} \phi ( [\mathbf{f}_j \|
\mathbf{p}_j - \mathbf{p}_{c_i} ]) \right)
\end{equation}
where $\gamma$ and $\phi$ are MLPs, $\mathcal{N}_{c_i}$ the k-nearest neighbors of point $\mathbf{p}_{c_i}$, $\mathbf{f}_j$ the latent features of point $\mathbf{p}_{j}$ and $\alpha_{ij}$ the attention coefficients between center $\mathbf{p}_{c_i}$ and point $\mathbf{p}_{j}$. The attention coefficients $\alpha_{ij}$ are computed using scaled dot-product, i.e. $\alpha_{ij} = $ softmax$\left(\frac{\theta_q(p_j)^T \theta_k(p_i)}{\sqrt{d}}\right)$, where $\theta_q, \theta_k$ are linear transformations: $\mathbb{R}^{3}\mapsto \mathbb{R}^{d}$.
\subsection{Loss Function}
\label{losses}
The selection of the loss function to be optimized is crucial for the task of simplification since we seek for a balance between the preservation of the structure and the salient features. A major barrier of most common distance metrics is the uniform weighting of points that can not reflect the perceptual differences between objects. As shown in many studies \citep{jin2020dr,li2019lbs,wen2019pixel2mesh++} the commonly used Chamfer distance (CD) between two point sets $\mathcal{P}_1,\mathcal{P}_2$ defined as: 
\begin{equation}
    d_{\mathcal{P}_1, \mathcal{P}_2} = \sum_{x \in \mathcal{P}_1} \min_{y \in \mathcal{P}_2} \| x-y \|^2 + \sum_{y \in \mathcal{P}_2} \min_{x \in \mathcal{P}_1} \|x-y \|^2 
    \label{eq:chamf}
\end{equation}
can only describe the overall surface structure similarity between the two sets without taking into account the high frequency details of each point cloud. Figure \ref{fig:smoothcd} illustrates an example of such case. 
Similarly, the point to surface distance between points of a set $\mathcal{P}$ and a surface $\mathcal{M}$ as well as the Hausdorff distance can not preserve salient points of the object rather than the global appearance. Several 3D perceptual metric studies \citep{lee2005mesh,lavoue2006perceptually,lavoue2009local,zhang2019feature} have pointed out that features such as curvature and roughness of a 3D model are highly correlated with the visual perception and should be maintained at the simplified point cloud. To train our model for the simplification task it is essential to devise a loss function that preserves both the salient features along with the structure of the point cloud.

\subsubsection{Adaptive Chamfer Distance}
As can be easily observed, the first term of eq. \eqref{eq:chamf} measures the preservation of the overall structure of $\mathcal{P}_1$ by $\mathcal{P}_2$, in a uniform way. To break the uniformity of the first term of CD we introduced a weighting factor $w_x$ in eq. \ref{eq:modify} that penalizes the distances between the two sets at the points with high salient features and ensures that they will be preserved at the simplified point cloud. We define the modified adaptive Chamfer distance as:  
\begin{equation}
    d^{Adapt}_{\mathcal{P}_1, \mathcal{P}_2} = \sum_{x \in \mathcal{P}_1} w_{\bar{\mathcal{K}}(x)} \min_{y \in \mathcal{P}_2} \| x-y \|^2 + \sum_{y \in \mathcal{P}_2} \min_{x \in \mathcal{P}_1} \|x-y \|^2    \label{eq:modify}
\end{equation}
where $\mathcal{P}_1$ denotes the initial point cloud, $\mathcal{P}_2$ the simplified one, and $w_{\bar{\mathcal{K}}(x)}$ a weighting factor proportional to the mean curvature $\bar{\mathcal{K}}$ at point $x$\footnote{We define the weights $w_x$ using the sigmoid of the normalized curvatures divided by a temperature scalar $\tau=10$ to amplify high curvature values.}. Since we only aim to retain salient points of $\mathcal{P}_1$, we avoid applying a similar weighting factor to the second term of eq.  \eqref{eq:chamf} to prevent the optimization process from getting trapped at local minima. 
\subsubsection{Curvature Preservation}
Additional to the adaptive CD, we make use of a loss term to reinforce the selection of high curvature points of the input point cloud. To quantify the preserved salient features of the input we define an error to measures the average point-wise curvature distance between the two point clouds:
\begin{equation}
    \mathcal{E}_c = \left( \frac{1}{\rvert\mathcal{P}_1\lvert}\sum_{ x \in \mathcal{P}_1} \| \bar{\mathcal{K}}_1(x) - \bar{\mathcal{K}}_2(\text{NN}(x,\mathcal{P}_2)) \| ^2 \right) ^{1/2}
    \label{eq:curv}
\end{equation}
where $\text{NN}(x,\mathcal{P}_2)$  the nearest neighbour of $x$ in set $\mathcal{P}_2$, and $\bar{\mathcal{K}}(\cdot)$ denotes the mean curvature. We refer to this error as Curvature Error (CE).  

\subsubsection{Overall Objective}
We used a combination of the two aforementioned losses as the total objective to be minimized: 
\begin{equation}
\mathcal{L}(\mathcal{P}_1 ,\mathcal{P}_2) =   d^{Adapt}_{\mathcal{P}_1, \mathcal{P}_2}+ \lambda \mathcal{E}_c 
\end{equation}
The first term ensures that the selected points cover the surface of the input, while the latter encourages the selection of high curvature points. 

\subsection{Implementation Details} 
We implemented the projector network using three multi-layer perceptrons (MLP) followed by Batch Normalization \citep{batchnorm2015} and ReLU activation functions \citep{nair2010rectified}. The filter sizes were set to 64. The GNN following the stacked MLPs was also ReLU activated with a filter size of 64. The initial point of FPS is randomly selected since we did not observe any influence on the performance. We selected 15 neighbours for each cluster center selected by FPS. The filter size of the attention-based refinement layer was set to 3, mapping the (64+3) features of the selected points to $(x,y,z)$ coordinates. We trained our model for 150 epochs with learning rate of 0.001 and a weight decay of 0.99 on every epoch using the Adam optimizer \citep{kingma2014adam}. 

\section{Evaluation Criteria}
To assess the performance of the simplified models generated by our method in terms of visual perception we define several metrics that measure the similarity between the two point cloud models. 

\subsection{Roughness Preservation}
Roughness describes the deviation of a point from the surface defined by its neighbours and has been identified as a salient feature in many visual perception studies \citep{lee2005mesh,wang2019fast}. Similar to the curvature preservation loss, we calculate the roughness preservation error by substituting the curvature values with roughness in eq \eqref{eq:curv}. We refer to this error as RE.  

\subsection{Point Cloud Structural Distortion Measure}
Additionally to curvature and roughness preservation metrics, we also calculated the structural similarity score between the two point clouds, that has shown to highly correlate with the human perception \citep{lavoue2006perceptually}. In particular, the Point-Cloud Structural Distortion Measure (SDM) can be defined as:
\begin{equation}
    D(\mathcal{P}_1, \mathcal{P}_2) = \frac{\alpha \mathcal{L}(p_i, \hat{p}_i)+\beta\mathcal{C}(p_i, \hat{p}_i) +\gamma \mathcal{S}(p_i, \hat{p}_i)}{\alpha + \beta + \gamma} 
\end{equation}

\begin{equation}
    \mathcal{L}(p_i, \hat{p}_i) = \frac{||\bar{\mathcal{K}_1}(p_i) -\bar{\mathcal{K}_2}(\hat{p}_i) ||} 
    {\max( \bar{\mathcal{K}_1}(p_i),\bar{\mathcal{K}_2}(\hat{p}_i))}
\end{equation}
\begin{equation}
     \mathcal{C}(p_i, \hat{p}_i)= \frac{||\sigma_{\bar{\mathcal{K}_1}}(p_i) -\sigma_{\bar{\mathcal{K}_2}}(\hat{p}_i) ||}
    {\max( \bar{\mathcal{K}_1}(p_i),\bar{\mathcal{K}_2}(\hat{p}_i))}
\end{equation}
\begin{equation}
     \mathcal{S}(p_i, \hat{p}_i)=  \frac{||\sigma_{\bar{\mathcal{K}_1}}(p_i)\sigma_{\bar{\mathcal{K}_2}}(\hat{p}_i) - \sigma_{\bar{\mathcal{K}}_{12}}(p_i,\hat{p}_i)^2 ||}
    {\sigma_{\bar{\mathcal{K}_1}}(p_i)\sigma_{\bar{\mathcal{K}_2}}(\hat{p}_i)}
\end{equation}
where $\mathcal{K}_1$, $\mathcal{K}_2, \sigma_{\bar{\mathcal{K}_1}}, \sigma_{\bar{\mathcal{K}_2}}, \sigma_{\bar{\mathcal{K}}_{12}}(p_i,\hat{p}_i)$ are the mean, the gaussian-weighted standard deviation and the covariance of the curvatures for point $p_{i}$ in  $\mathcal{P}_1$ and its corresponding point $\hat{p}_i$ in $\mathcal{P}_2$, respectively. We establish the correspondence between the two point clouds using the 1-nearest neighbor for each point. The global similarity score is obtained using \emph{Minkowski pooling} as suggested in \citep{lavoue2011multiscale}.

\subsection{Normals Consistency}
Point normals are highly related to visual appearance and could be indicators of sharp and smooth areas. To measure the consistency of normals' orientations between the two models we use the bi-directional cosine similarity as: 

\begin{equation}
\begin{split}
\mathcal{E}_n =\frac{1}{\rvert\mathcal{P}_1\lvert}\sum_{\substack{ x \in \mathcal{P}_1 \\ y \in NN(x,\mathcal{P}_2)}}1 - {\mathbf {n_x} \cdot \mathbf {n_y}  \over \|\mathbf {n_x} \|\|\mathbf {n_y} \|} + \\
\frac{1}{\rvert\mathcal{P}_2\lvert}\sum_{\substack{ y \in \mathcal{P}_2 \\ x \in NN(y,\mathcal{P}_1)}}1 - {\mathbf {n_x} \cdot \mathbf {n_y}  \over \|\mathbf {n_x} \|\|\mathbf {n_y} \|}
\end{split}
\end{equation}
where $\mathbf {n_x}$ denotes the normal at point $x$ and $NN(x,\mathcal{P}_2)$ the nearest neighbour of $x$ in set $\mathcal{P}_2$, calculated as described in Section \ref{sec:normalest}.

\begin{figure*}[!ht]
\begin{center}

  \includegraphics[width=0.8\linewidth]{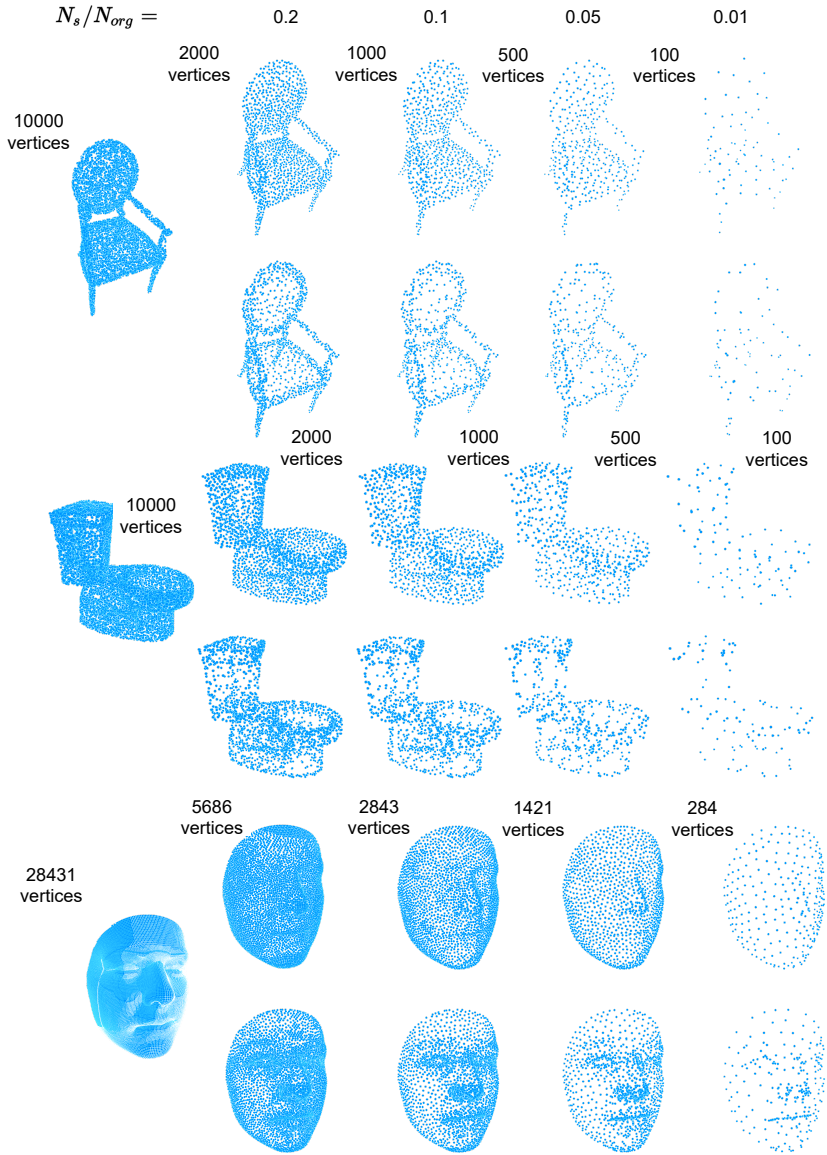}
\end{center}
  \caption{Qualitative comparison between FPS (top row) and the proposed (bottom row) methods, at different simplification ratios. Differences between
the two methods can be found at coarse and smooth areas, where the proposed model favours the preservation of high-frequency details of the input point cloud. Notice that high curvature areas such as the back of the chair, the rim of the toilet and the eyes and mouth of the face point clouds are preserved using the proposed method in contrast to the smooth results that FPS method produces.}
\label{fig:example}
\end{figure*}

\begin{figure*}[!ht]
\begin{center}
   \includegraphics[width=0.75\linewidth]{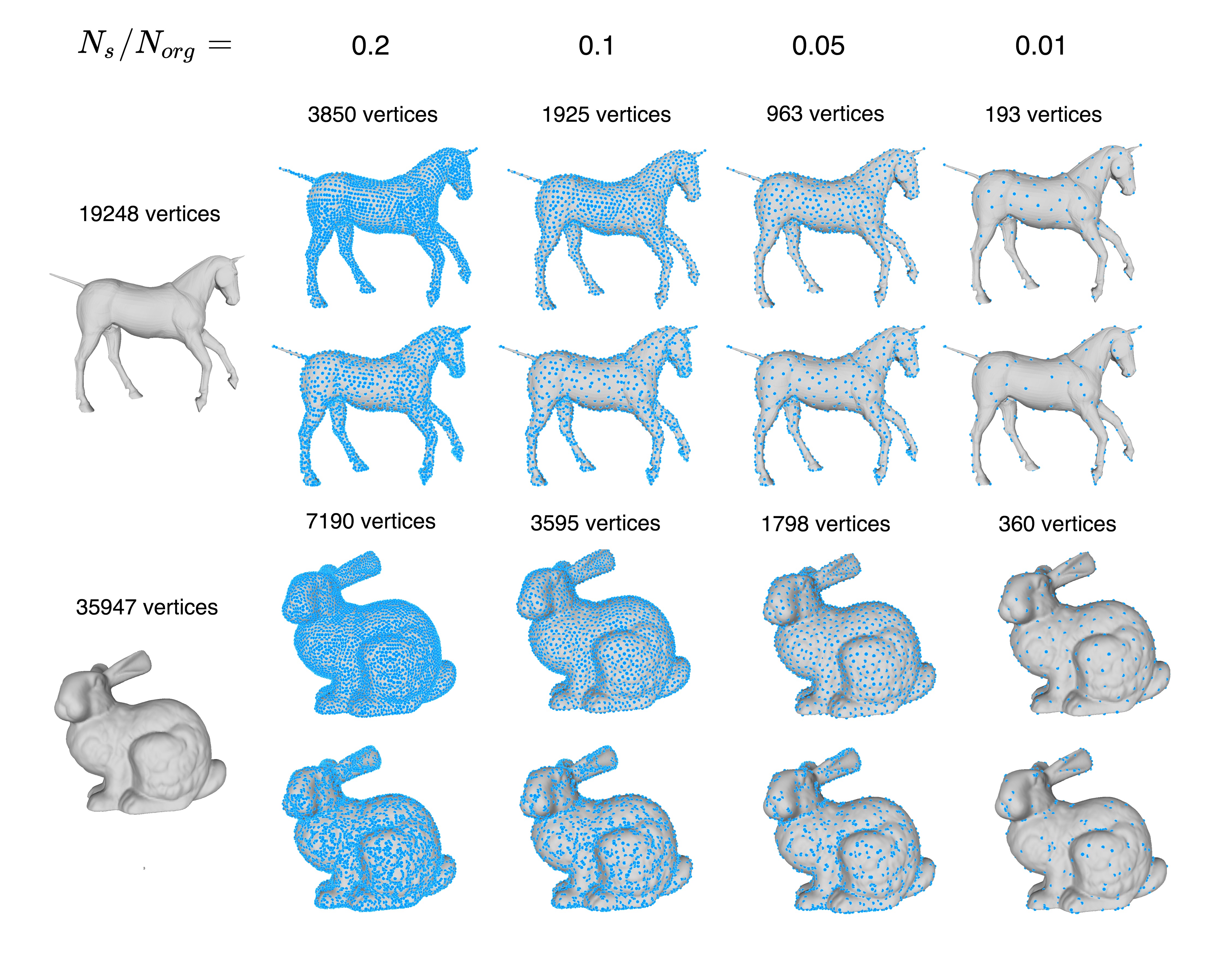}
\end{center}
   \caption{Qualitative comparison between QEM (top row) and the proposed (bottom row) methods, at different simplification ratios. Point clouds are rendered on top of the original mesh surfaces to better visualize high-curvature areas. Notice that the proposed method favours the selection of points at rough areas such as legs, head and ears. Figure better viewed in zoom. }
\label{fig:examples}
\end{figure*}

\section{Experiments}
Throughout this section we extensively evaluated the proposed method with both quantitative and qualitative experiments. We compare our approatch against the following baselines: uniform subsampling, farthest point sampling, quadric error metric (QEM) simplification  \citep{garland1997surface} along with a top curvature points sampling (TCP) where the top-k curvature points are selected from the input point cloud.

\subsection{Datasets} 
To evaluate our method we used several publicly available 3D datasets, with different characteristics.
 The TOSCA \citep{bronstein2008numerical} dataset consists of 80 synthetic 
 meshes with 9 different deformable objects. TOSCA is a standard benchmark for the evaluation of simplification methods since it includes high-resolution meshes with a varying number of vertices between 10K to 50K. In addition, it is an excellent candidate to assess feature-preserving simplification, since most of its meshes are non-smooth consisting of high curvature regions. We split TOSCA dataset into 80\% training and 20\% testing, in a stratified manner. Additionally, we used the popular ModelNet10 dataset \citep{wu20153d} which was initially preprocessed to remove redundant points and retain only high-resolution meshes that have more than 2K vertices. For ModelNet10 we used the official train-test split. Finally, we assessed the performance of our method in a fixed topology setting, i.e. where all meshes have the same connectivity, as in MeIn3D face dataset  \citep{booth2018large}, where the surfaces of which are significantly smoother than the previous use cases. This experiment aims to assess the ability of the proposed method to simplify fixed topology meshes by fine-tuning the model with only a few samples from the dataset. MeIn3D dataset consists of 10K meshes with 28431 vertices each, formed from distinct human faces. For MeIn3D experiments, we randomly selected 10\%  of the meshes for training and 90\% for testing.

\begin{figure}[!ht]
\begin{center}

   \includegraphics[width=\linewidth]{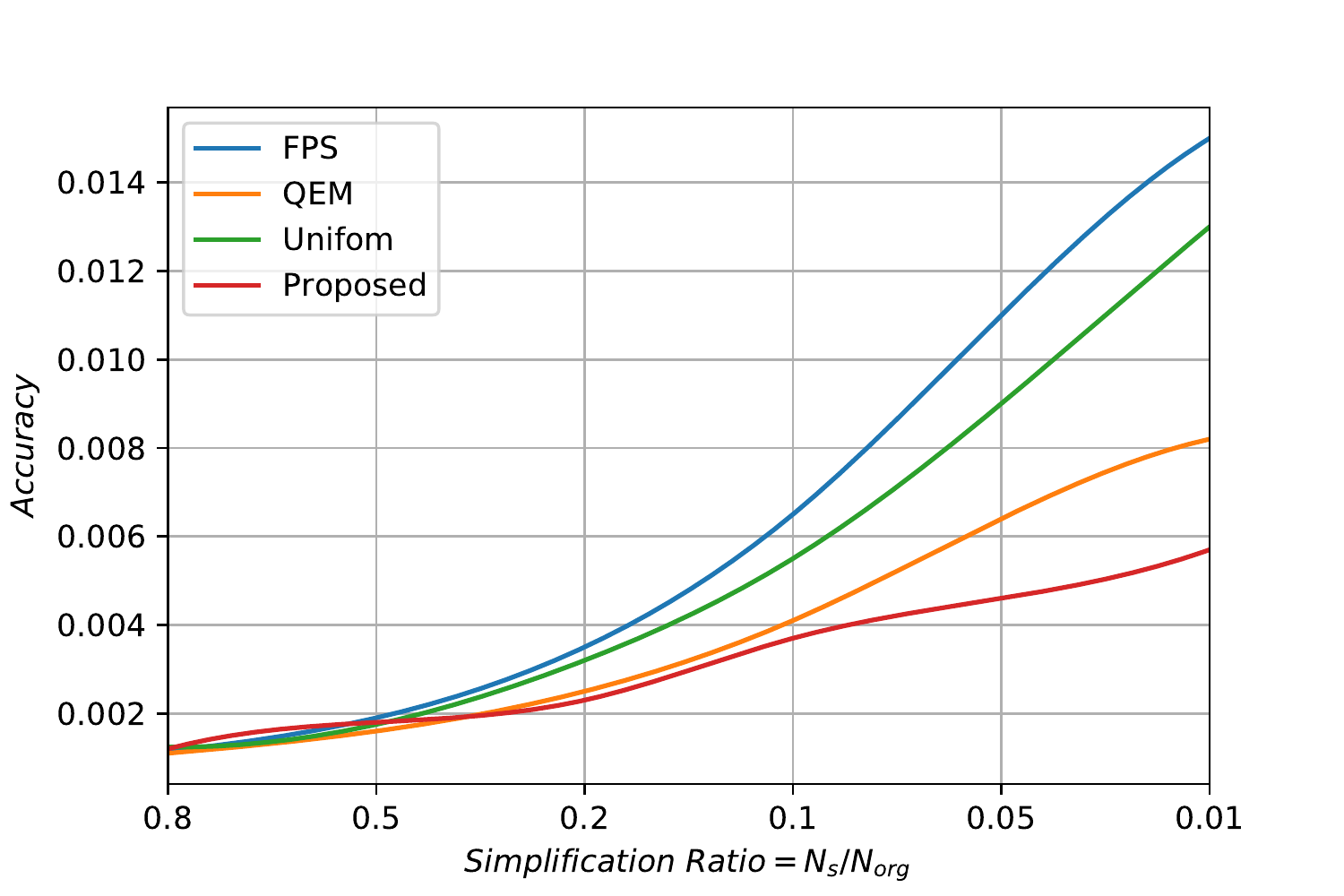}
\end{center}
   \caption{Curvature preservation error comparison for the TOSCA dataset. Quantitative evaluation of different point cloud simplification methods at different simplification ratios.}
\label{fig:curv}
\end{figure}

\begin{table*}
\centering

\resizebox{\linewidth}{!}{
\begin{tabular}{l|cccc|cccc|cccc|}
                  & \multicolumn{12}{c}{TOSCA}                                                   \\
                  & \multicolumn{4}{c}{$N_s/N_{org}=$ 0.8}                            & \multicolumn{4}{c}{$N_s/N_{org}=$ 0.5}                            & \multicolumn{4}{c|}{$N_s/N_{org}=$ 0.3}                                           \\
Method            & CD                             & NC                                    & RE($\times 10^{-4}$)                          & \multicolumn{1}{c|}{MSDM($\times 10^{-3}$)}                           & CD          & NC                                    & RE($\times 10^{-4}$)                          & \multicolumn{1}{c|}{MSDM($\times 10^{-3}$)}                           & CD & NC                                    & RE($\times 10^{-4}$)                          & \multicolumn{1}{c|}{MSDM($\times 10^{-3}$)}                           \\ \hline
Random            & 0.14                                 & \textbf{0.093}                        & 2.87                                 & \multicolumn{1}{c|}{2.43}                                 & 0.49                                & \textbf{0.106}                        & 3.04                                 & \multicolumn{1}{c|}{3.03}                                 & 1.04    & \textbf{0.225}                        & 3.65                                 & \multicolumn{1}{c|}{4.33}                                 \\
TCP               & 1.11                                & 0.147                                 & 2.86                                 & \multicolumn{1}{c|}{2.61}                                 & 10.0                              & 0.272                                 & 3.36                                 & \multicolumn{1}{c|}{4.05}                                 & 30.8   & 0.357                                 & 4.19                                 & \multicolumn{1}{c|}{6.57}                                 \\

FPS               & 0.09                                 & 0.103                                 & \textcolor{red}{ \textbf{2.85}} & \multicolumn{1}{c|}{2.32}                                 & 0.29                              & 0.245                                 & \textcolor{red}{ \textbf{2.97}} & \multicolumn{1}{c|}{2.92}                                 & 0.67     & 0.255                                 & \textcolor{red}{ \textbf{3.52}} & \multicolumn{1}{c|}{4.22}                                 \\
QEM               & 0.09                               & 0.103                                 & \textbf{2.81}                        & \multicolumn{1}{c|}{2.33}                                 & 0.29                              & 0.214                                 & \textbf{2.96}                        & \multicolumn{1}{c|}{2.91}                                 & 0.84     & \textcolor{red}{ \textbf{0.248}} & 3.54                                 & \multicolumn{1}{c|}{4.27}                                 \\ \hline
Proposed-MeIn3D   & 0.05                               & 0.104                                 & 2.88                                 & \multicolumn{1}{c|}{2.30}                                 & 0.25                                & 0.244                                 & 3.06                                 & \multicolumn{1}{c|}{2.87}                                 & 0.65     & 0.255                                 & 3.54                                 & \multicolumn{1}{c|}{\textcolor{red}{ \textbf{4.07}}} \\
Proposed-ModelNet & \textcolor{red}{ \textbf{0.03}} & 0.103                                 & 2.87                                 & \multicolumn{1}{c|}{\textcolor{red}{ \textbf{2.29}}} & \textcolor{red}{ \textbf{0.23}} & 0.211                                 & 3.05                                 & \multicolumn{1}{c|}{\textcolor{red}{ \textbf{2.83}}} & \textcolor{red}{\textbf{0.64}}     & 0.259                                 & \textbf{3.51}                        & \multicolumn{1}{c|}{4.08}                                 \\
Proposed-TOSCA    & \textbf{0.03}                        & \textcolor{red}{ \textbf{0.102}} & 2.86                                 & \multicolumn{1}{c|}{\textbf{2.21}}                        & \textbf{0.23}                        & \textcolor{red}{ \textbf{0.193}} & 3.03                                 & \multicolumn{1}{c|}{\textbf{2.79}}                        & \textbf{0.63}     & 0.254                                 & 3.55                                 & \multicolumn{1}{c|}{\textbf{4.04}}                        \\ \hline \hline 
     & \multicolumn{4}{c|}{ }                                                                                                & \multicolumn{4}{c|}{ }                                                                                                  & \multicolumn{4}{c|}{ }

\\
                  & \multicolumn{4}{c|}{$N_s/N_{org}=$ 0.2 }                                                                                                & \multicolumn{4}{c|}{$N_s/N_{org}=$0.1 }                                                                                                  & \multicolumn{4}{c|}{$N_s/N_{org}=$0.05 }                                                                                                \\ 
\hline
Uniform            & 1.63                            & 0.312                           & 4.45                            & 6.07                            & 3.35                            & 0.342                            & 4.91                            & 10.7                            & 6.68                            & 0.369                           & 5.71                            & 19.2                             \\
TCP               & 51.3                            & 0.625                           & 4.99                            & 9.52                            & 129.4                           & 0.732                            & 6.42                            & 17.8                            & 172.5                           & 0.793                           & 6.20                            & 32.4                             \\
FPS               & \textbf{0.81}                   & 0.307                           & 4.71                            & \textcolor{red}{\textbf{5.13}}  & \textbf{1.93}                   & 0.341                            & 4.82                            & 9.64                            & \textbf{3.94}                   & \textbf{0.321}                  & 5.56                            & 18.3                             \\
QEM               & 1.35                            & \textbf{\textcolor{red}{0.291}} & \textcolor{red}{\textbf{4.01}}  & 5.36                            & 2.64                            & \textcolor{red}{\textbf{0.310}}  & 4.79                            & 10.4                            & \textcolor{red}{\textbf{4.77}}  & 0.338                          & 5.53                            & 18.4                             \\ 
\cite{LiuSpectral}  & 2.17 &  0.358 & 4.39 & \multicolumn{1}{c|}{5.39} & 3.12 &  0.331 & 4.96 & \multicolumn{1}{c|}{10.4} & 5.62 &  0.441 &  5.96 & \multicolumn{1}{c|}{18.5} \\

\hline

Proposed MeIn3D   & 1.14                            & 0.293                           & 4.15                            & 5.64                            & 2.53                            & 0.313                           & \textcolor{red}{\textbf{4.47}}  & \textcolor{red}{\textbf{8.15}}  & 5.36                            & 0.364                           & 5.01                            & 17.7                             \\
Proposed ModelNet & 1.15                            & 0.310                           & \textcolor{red}{\textbf{4.01}}  & 5.53                            & 2.51                            & 0.312                            & 4.81                            & 9.72                            & 5.19                            & 0.341                          & \textcolor{red}{\textbf{4.99}}  & \textcolor{red}{\textbf{17.4}}   \\
Proposed TOSCA    & \textcolor{red}{\textbf{1.12}}  & \textbf{0.290}                  & \textbf{3.91}                   & \textbf{5.01}                   & \textcolor{red}{\textbf{2.45}}  & \textbf{0.307}                   & \textbf{4.41}                   & \textbf{7.84}                   & 4.93                            & \textbf{\textcolor{red}{0.333}} & \textbf{4.93}                   & \textbf{16.5}                    \\
\hline
\end{tabular}}
\caption{Simplification performance tested on TOSCA dataset. Best approaches highlighted are highlighted in \textbf{bold} and second best in \textcolor{red}{\textbf{red}}. We refer to the dataset used for training as ``Proposed-\textit{Dataset}"}
\label{tab:tosca}
\end{table*}

\begin{table*}[]
\centering
\resizebox{\linewidth}{!}{
\begin{tabular}{l|cccc|cccc|cccc|}
                  & \multicolumn{12}{c}{ModelNet}                                                                                                \\
                  & \multicolumn{4}{c}{$N_s/N_{org}=$ 0.8}                                                                                                                                          & \multicolumn{4}{c}{$N_s/N_{org}=$0.5}                                                                                     & \multicolumn{4}{c|}{$N_s/N_{org}=$0.3}                                                           \\
Method            & CD($\times 10^{-4}$)   & NC                                    & RE($\times 10^{-4}$)                          & \multicolumn{1}{c|}{MSDM($\times 10^{-3}$)}                           & CD($\times 10^{-4}$)      & NC             & RE($\times 10^{-4}$)                          & \multicolumn{1}{c|}{MSDM($\times 10^{-3}$)}                           & CD($\times 10^{-4}$)       & NC             & RE($\times 10^{-4}$) & \multicolumn{1}{c|}{MSDM($\times 10^{-3}$)}                           \\ \hline
Random            & 1.74                                 & \textbf{0.181}                        & 4.91                                 & \multicolumn{1}{c|}{1.14}                                 & 3.13  & \textbf{0.201} & 5.16                                 & \multicolumn{1}{c|}{1.53}                                 & 6.01   & \textbf{0.333} & 5.37        & \multicolumn{1}{c|}{1.99}                                 \\
CP                & 14.01                                & 0.288                                 & 5.01                               & \multicolumn{1}{c|}{1.12}                                 & 55.12 & 0.371          & 5.98                                 & \multicolumn{1}{c|}{1.68}                                 & 117.11 & 0.527          & 6.63        & \multicolumn{1}{c|}{2.71}                                 \\
FPS               & \textbf{0.89}                        & \textcolor{red}{\textbf{0.195}} & \textcolor{red}{\textbf{4.71}} & \multicolumn{1}{c|}{\textcolor{red}{\textbf{1.01}}} & \textcolor{red}{\textbf{1.93}}  & \textcolor{red}{\textbf{0.213}}          & 4.89                                 & \multicolumn{1}{c|}{\textcolor{red}{\textbf{1.35}}} & \textcolor{red}{\textbf{3.02}}   & \textcolor{red}{\textbf{0.352}}         & 5.57        & \multicolumn{1}{c|}{2.08} \\
QEM               & 1.35                                 & 0.211                                 & 4.98                                 & \multicolumn{1}{c|}{1.14}                                 & 2.84  & 0.224          & 5.12                                 & \multicolumn{1}{c|}{1.48}                                 & 3.05   & 0.382          & 5.57        & \multicolumn{1}{c|}{2.44}                                 \\ \hline
Proposed-MeIn3D   & 2.32                                & 0.353                                 & 5.12                                 & \multicolumn{1}{c|}{1.11}                                 & 2.81 & 0.365          & 5.23                                 & \multicolumn{1}{c|}{1.50}                                 & 3.72  & 0.473          & 5.53        & \multicolumn{1}{c|}{2.15}                                 \\
Proposed-ModelNet & \textcolor{red}{\textbf{0.91}} & 0.207                                 & \textbf{4.61}                        & \multicolumn{1}{c|}{\textbf{0.99}}                        & \textbf{1.12}  & 0.216          & \textbf{4.72}                        & \multicolumn{1}{c|}{\textbf{1.28}}                        & \textbf{2.74}   & 0.371          & \textbf{5.01}        & \multicolumn{1}{c|}{\textbf{1.87}}                        \\
Proposed TOSCA    & 2.12                                 & 0.270                                 & 4.82                                 & \multicolumn{1}{c|}{1.07}                                 & 2.98  & 0.283          & \textcolor{red}{\textbf{4.86}} & \multicolumn{1}{c|}{1.42}                                 & 4.11   & 0.401          & \textcolor{red}{\textbf{5.26}}        & \multicolumn{1}{c|}{\textcolor{red}{\textbf{2.03}}}                                 \\ \hline \hline    & \multicolumn{4}{c|}{ }                                                                                                & \multicolumn{4}{c|}{ }                                                                                                  & \multicolumn{4}{c|}{ }

                                                            \\
                                                            
                & \multicolumn{4}{c|}{$N_s/N_{org}  = $ 0.2}             & \multicolumn{4}{c|}{$N_s/N_{org} = $ 0.1}             & \multicolumn{4}{c|}{$N_s/N_{org}= $ 0.05}             \\
\hline
Uniform         & 8.01          & 0.568          & 5.91          & 2.83          & 20.4          & 0.655          & 6.19          & 4.92          & 41.02          & 0.793          & 6.57          & 8.19          \\
TCP              & 197.3         & 0.898          & 7.25          & 3.87          & 403.1         & 0.937          & 7.84          & 7.11          & 611.6          & 0.952          & 7.01          & 12.81         \\
FPS             & \textbf{3.12}          & \textbf{0.505} & 6.05          & 2.74          & \textbf{7.56} & 0.641          & 6.39          & 4.81          & \textbf{16.01} & 0.744          & 6.48          & 8.38          \\
QEM             & 3.45          & \textcolor{red}{\textbf{0.513}}          & 5.94         & 3.01          & 9.45          & 0.625          & 6.13          & 5.19          & 21.43          & 0.724          & 6.25          & 9.12          \\ \hline

Proposed-MeIn3D & 4.02         & 0.531          & 5.93          & 2.86          & 29.31         & 0.610          & 6.08          & 4.76          & 45.12          & 0.701          & 6.33          & 8.02          \\
Proposed-ModelNet       & \textcolor{red}{\textbf{3.32}} & 0.515          & \textcolor{red}{\textbf{5.79}}          & \textbf{2.68} & \textcolor{red}{\textbf{8.24}}          & \textcolor{red}{\textbf{0.606}}          & \textcolor{red}{\textbf{6.06}}          & \textbf{4.61} & \textcolor{red}{\textbf{17.24}}          & \textcolor{red}{\textbf{0.696}}          & \textcolor{red}{\textbf{6.25}}          & \textbf{7.92}       \\
Proposed-TOSCA  & 4.35          & 0.523          & \textbf{5.77} & \textcolor{red}{\textbf{2.72}}          & 9.42          & \textbf{0.603} & \textbf{5.91} & \textcolor{red}{\textbf{4.64}}          & 22.18          & \textbf{0.688} & \textbf{6.04} & \textcolor{red}{\textbf{7.96}}  \\\hline
\end{tabular}}

\caption{Simplification performance tested on ModelNet10 dataset.}
\label{tab:modelnet}
\end{table*}

\begin{table*}[!h]
\centering
\resizebox{\linewidth}{!}{
\begin{tabular}{l|cccc|cccc|cccc|}
                  & \multicolumn{12}{c}{MeIn3D}                                                                                                                                                                                                                      \\
                  & \multicolumn{4}{c}{$N_s/N_{org}=$ 0.8}                                                                                              & \multicolumn{4}{c}{$N_s/N_{org}=$0.5}                                                                                               & \multicolumn{4}{c|}{$N_s/N_{org}=0.3$}                                                                                     \\
Method            & CD($\times 10^{-4}$)           & NC                               & RE($\times 10^{-4}$)           & MSDM($\times 10^{-3}$)         & CD($\times 10^{-4}$)           & NC                               & RE($\times 10^{-4}$)           & MSDM($\times 10^{-3}$)         & CD($\times 10^{-4}$) & NC                               & RE($\times 10^{-4}$)           & MSDM($\times 10^{-3}$)          \\ 
\hline
Random            & 0.74                           & 0.1083                           & 2.46                           & 0.99                           & 1.12                           & 0.120                           & 2.74                           & 1.24                           & 1.26                 & 0.169                           & 3.43                           & 2.31                            \\
TCP               & 12.35                          & 0.2114                           & 2.41                           & 0.97                           & 43.06                          & 0.3272                           & \textbf{\textcolor{red}{2.52}} & 1.54                           & 89.24                & 0.5711                           & \textbf{2.89}                  & 2.91                            \\
FPS               & \textbf{0.59}                  & \textbf{0.103}                  & \textcolor{red}{\textbf{2.32}} & \textbf{0.86}                  & \textbf{0.97}                  & \textcolor{red}{\textbf{0.105}} & 2.53                           & \textcolor{red}{\textbf{1.15}} & \textbf{1.05}        & \textbf{\textcolor{red}{0.108}} & 3.21                           & 2.28                            \\
QEM               & 0.94                           & 0.112                           & 2.52                           & 1.06                           & 1.36                           & 0.139                           & 2.76                           & 1.44                           & 1.94                 & 0.150                           & 3.53                           & 2.54                            \\ 
\hline
Proposed MeIn3D   & \textbf{\textcolor{red}{0.61}} & \textbf{\textcolor{red}{0.104}} & \textbf{2.29}                  & \textbf{\textcolor{red}{0.90}} & \textcolor{red}{\textbf{0.98}} & \textbf{0.105}                  & \textbf{2.46}                  & \textbf{1.07}                  & \textbf{\textcolor{red}{1.15}}                 & \textbf{0.105}                  & \textbf{2.89}                  & \textbf{1.76}                   \\
Proposed ModelNet & 1.15                           & 0.111                           & 2.41                           & 1.02                           & 1.28                           & 0.123                           & 2.68                           & 1.43                           & 1.59                 & 0.165                           & 2.99                           & 2.09                            \\
Proposed TOSCA    & 1.04                           & 0.106                           & 2.41                           & 1.08                           & 1.21                           & 0.1171                           & 2.63                           & 1.33                           & 1.41                 & \textbf{\textcolor{red}{0.149}} & \textbf{\textcolor{red}{2.96}} & \textbf{\textcolor{red}{1.85}}  \\ \hline \hline  & \multicolumn{4}{c|}{ }                                                                                                & \multicolumn{4}{c|}{ }                                                                                                  & \multicolumn{4}{c|}{ }

                                                \\
                & \multicolumn{4}{c|}{$N_s/N_{org}  = $ 0.2}             & \multicolumn{4}{c|}{$N_s/N_{org} = $ 0.1}             & \multicolumn{4}{c|}{$N_s/N_{org}= $ 0.05}             \\
\hline
Uniform           & 1.42          & 0.198          & 4.15          & 2.81          & 3.46          & 0.313          & 6.73          & 5.92          & 5.52          & 0.481          & 7.05        & 12.4          \\
TCP                & 158.3         & 0.801          & 3.46          & 3.73          & 421.1         & 0.910          & 6.02          & 7.38          & 556.0         & 0.934          & 11.87       & 14.2          \\
FPS               & \textbf{1.12} & \textbf{0.121} & 3.64          & 2.96          & \textbf{1.93} & 0.195          & 6.29          & 5.98          & \textcolor{red}{\textbf{3.45}}          & 0.484          & 7.43        & 11.8          \\
QEM               & 2.01          & 0.185          & 4.53          & 3.01          & 2.52          & 0.198          & 6.31          & 5.71          & 3.65          & \textcolor{red}{\textbf{0.331}}          & 8.13        & 11.3          \\ \hline
Proposed-MeIn3D   & \textcolor{red}{\textbf{1.24}}          & \textcolor{red}{\textbf{0.128}}          & \textbf{3.15} & \textbf{2.30} & \textcolor{red}{\textbf{2.01}}          & \textbf{0.192} & \textbf{5.69} & \textbf{4.91} & \textbf{3.25} & \textbf{0.305} & \textbf{6.47}        & \textbf{10.6} \\
Proposed-ModelNet & 1.75          & 0.189          & 3.65          & 2.45          & 3.23          & 0.196          & \textcolor{red}{\textbf{5.73}}          & 5.10          & 4.02          & 0.369          & 7.02        & 10.9          \\ 
Proposed-TOSCA    & 1.54         & 0.168          & \textcolor{red}{\textbf{3.29}}          & \textcolor{red}{\textbf{2.41}}          & 2.32          & \textcolor{red}{\textbf{0.194}}          & 5.98          & \textcolor{red}{\textbf{5.06}}          & 3.82          & 0.342          &\textcolor{red}{\textbf{6.49}}        & \textcolor{red}{\textbf{10.8}}          \\\hline
\end{tabular}}

\caption{Simplification performance tested on MeIn3D dataset.}
\label{tab:mein3d}
\end{table*}

\subsection{Point Cloud Simplification}

\begin{figure*}[!ht]
\begin{center}

   \includegraphics[width=0.85\linewidth]{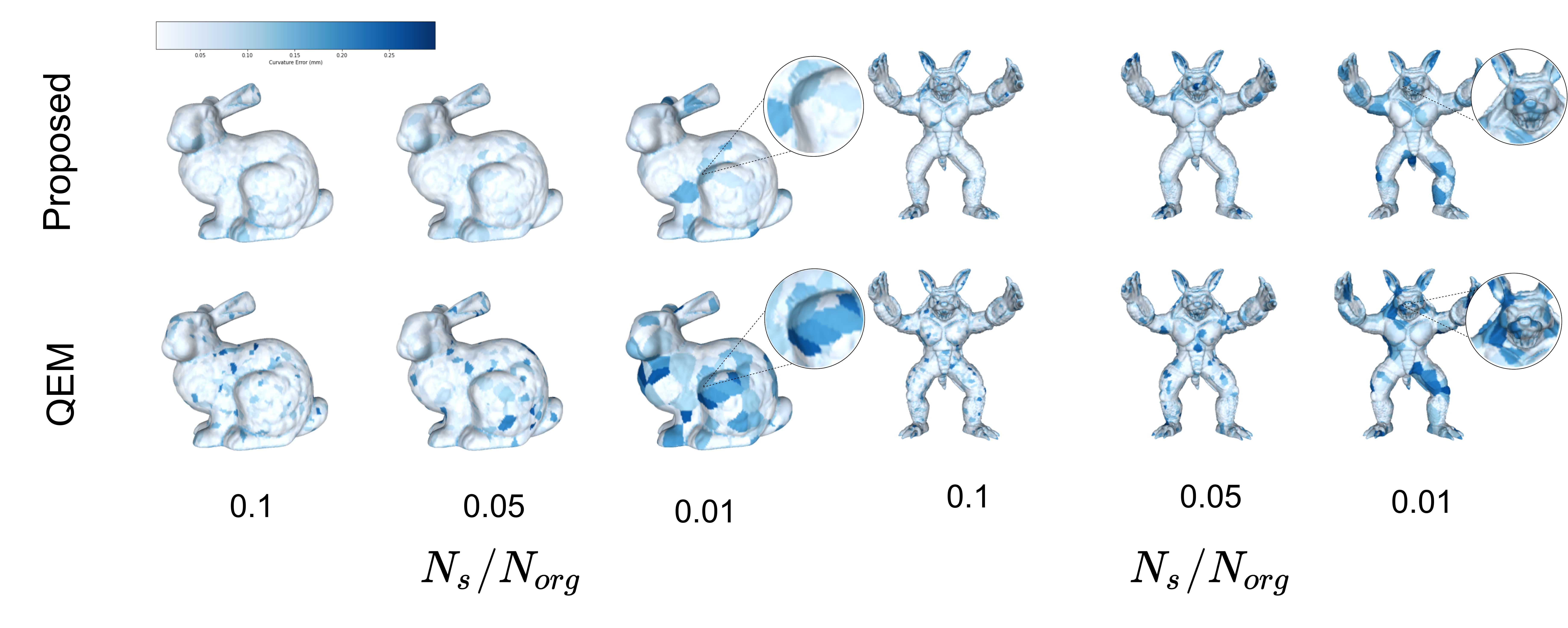}
\end{center}
   \caption{Curvature preservation error comparison for the TOSCA dataset. Comparison between different point cloud simplification methods at different simplification ratios. }
\label{fig:colorcoding}
\end{figure*}

In this section, we showcase the simplification performance of the proposed method. In Figure \ref{fig:example}, we visualize the differences between the FPS method that samples points from  the xyz-coordinate space and the proposed method that utilizes FPS to cluster the latent space. FPS samples almost uniformly the flat areas, in contrast with the proposed method that selects more points in areas with high curvature. 

Figure \ref{fig:examples} shows simplified point clouds at different scales, comparing QEM and the proposed method, where the point clouds are visualized on top of the mesh surfaces to highlight the salient regions. The proposed method favours point selection at the horse's nape and face in contrast to points at smooth areas, such as the thigh, to preserve salient features of the input point cloud. Intuitively, smooth areas require only a few points to describe their associated planes in contrast with coarse areas that demand many points in order to preserve their curvature. We selected to report and show the results for small simplification ratio values since we observed that the simplification error increases exponentially at low resolutions. This is due to the fact that small number of points cannot properly preserve both shape and details of the object.  We have also indicated that for high resolution meshes the perceptual quality is fairly preserved at high simplification ration (i.e. over 0.3). This can be also contended in Figure \ref{fig:curv}, where all models achieve similar curvature error for simplifications over 0.3. In contrast, baseline models fail to preserve salient points at small simplification ratios while the proposed method scales linearly. Moreover, as shown in Figure \ref{fig:colorcoding}, the QEM method performs poorly at coarse areas with increased curvature, whereas the proposed method achieves remarkable results even when only 1\% of the input points were retained. It is important to note that none of the topologies presented at Figures \ref{fig:examples}, \ref{fig:colorcoding} was part of the training set. For each dataset, we report both structural (i.e. Chamfer distance, normals consistency) as well as perceptual metrics (i.e. curvature and roughness preservation, point-cloud structural distortion measure) for the proposed and the baseline methods. Tables \ref{tab:tosca}, \ref{tab:modelnet}, \ref{tab:mein3d} indicate the superiority of the proposed method to maintain perceptual features of the input without sacrificing the overall structure of the shape at six indicative simplification ratios. In particular, in contrast to TPC method where only high curvature points were selected leading to an increased Chamfer distance, the proposed method manages a fair balance between structure and saliency. 

Although several recent methods have been proposed for mesh simplification \citep{nasikun2018fast,LiuSpectral,LescoatSpectral}, they rely on the eigendecomposition of the Laplacian matrix. This entails an overwhelmingly large processing run-time and memory consumption, which makes them impractical for large point clouds. In particular, for a mesh with $\approx$15K points,  \cite{nasikun2018fast} runs out of memory, while \cite{LiuSpectral} requires around 15min to execute, which makes them impractical. Nevertheless, we've managed to simplify the test set of TOSCA dataset using the method of \cite{LiuSpectral} for small simplification ratios (i.e. 0.2, 0.1 and 0.05). Results, included in Table \ref{tab:tosca}, indicate that the proposed model outperforms the recent spectral method under all metrics. 

We also experimented with a cross-dataset generalization scenario where different datasets were used for training and testing the model. Interestingly, it is observed that the proposed model can generalize well to out-of-distribution shapes and topologies and can be applied directly to any point cloud without fine-tuning. We argue that this is due to the diversity of shapes and the presence of many rough regions at the training sets of TOSCA and ModelNet that enforce the model to favour salient features. Experimental results also indicate that zero-shot scenario can be less successful when the model is pretrained using smooth non-diverse shapes such as those in MeIn3D.  
Thus, we believe that training the proposed model with an arbitrary topology dataset with coarse shapes such as TOSCA, can achieve remarkable results to out-of-distribution samples, as shown in Table \ref{tab:modelnet}.

\subsection{Classification of simplified point clouds}
\begin{figure}[!h]
\begin{center}
   \includegraphics[width=\linewidth]{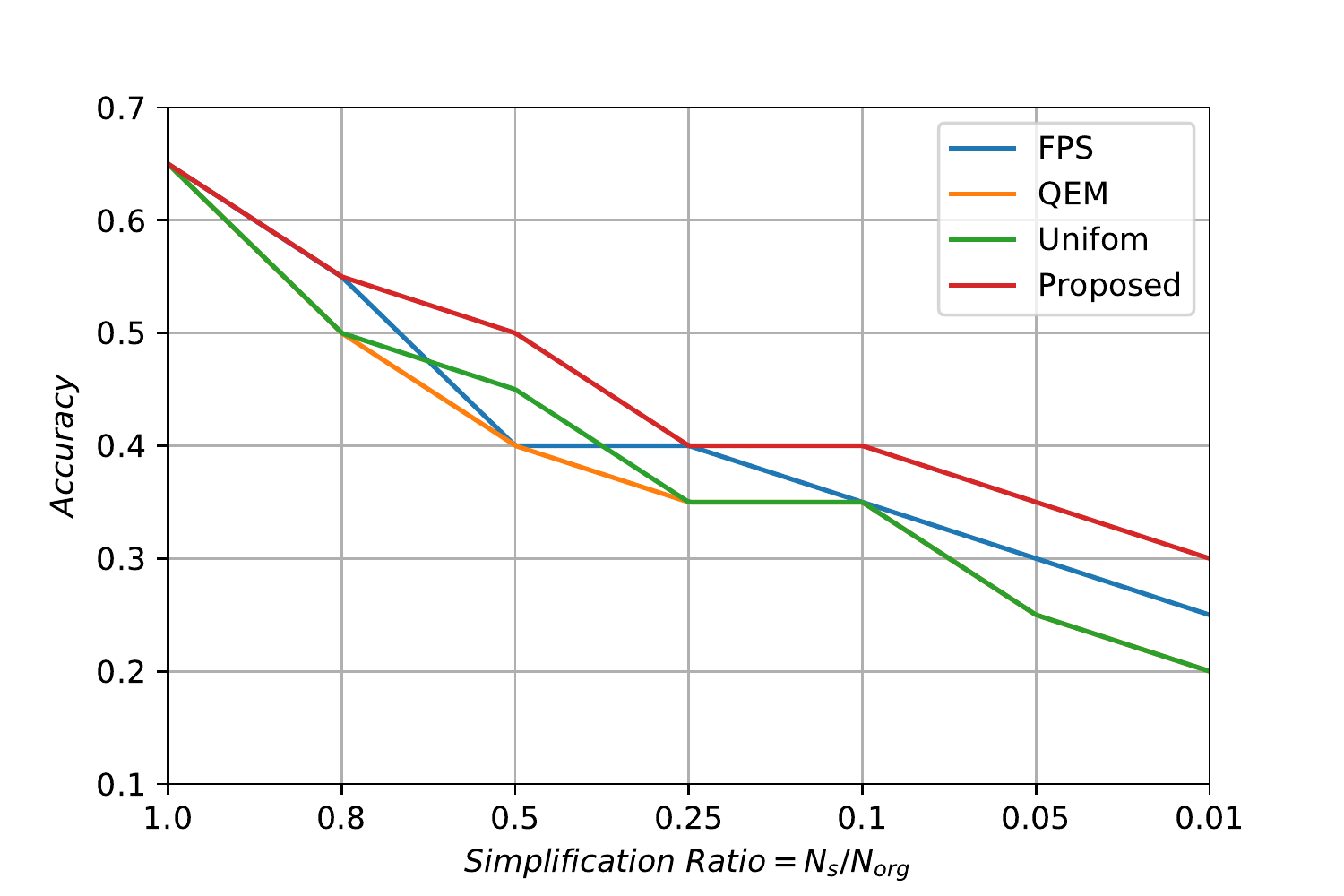}
\end{center}
   \caption{Classification accuracy of the pretrained PointNet++ on simplified point clouds at different ratios.}
\label{fig:classification}
\end{figure}

To further assess the simplification quality, we used a pretrained shape classification model and measured its classification accuracy on the simplified point clouds. In particular, we trained a PointNet \citep{qi2017pointnet} model on the train split of TOSCA dataset. We used the proposed and the baseline methods to simplify the remaining test split. In figure \ref{fig:classification} we show the classification performance, in terms of accuracy, of the compared methods at different simplification ratios. 
It is important to note that the scope of this experiment is to demonstrate that the simplified point clouds produced by the proposed method can be better classified, by a pretrained classifier, compared to the ones produced by the baselines. Boosting the point cloud classification performance remains out of the scope of this paper since category indicative points are not always correlated with the visual appearance of the model.  We report results of the original test set performance at simplification ratio equal to 1. It can be easily seen that the proposed model degrades with a smaller slope at extreme simplification ratios, compared to the baseline models. We argue that the performance drop of the baseline models could be attributed to the uniform way of sampling points that may drive to decimation of salient points that characterize the point cloud. In contrast, the perceptually influenced simplification of the proposed method selects and rearranges points according to their visual importance, ensuring that the salient features will be decimated last.    

\subsection{User-Study}
Apart from the error measures used to quantitatively assess the proposed method, we also performed a user study to quantify the ability of the proposed method to select points that correlate with human perception. In particular, we adopted the paired comparison approach where each
human subject was shown a reference point cloud, and two simplified point clouds, one by our method and one by a competing method. Users were asked to evaluate the two simplified point clouds in terms of the overall shape and identity similarity and select the one that preserves most of the perceptual details of the reference one. In total, forty users were asked to evaluate eighteen point cloud pairs, at different simplification ratios. Results are reported in Table \ref{tab:user}. In average, the users selected 14 out of the 18 point clouds produced by the proposed method, as the ones preserving most of the visual features. 
\begin{table}[!ht]
\centering
\begin{tabular}{|l|c|} 
\hline
Method & User preference (\%) \\
 \hline
 Proposed vs QEM    & \textbf{0.73}/0.28\\ 
 Proposed vs Random &  \textbf{0.78}/0.22 \\ 
 Proposed vs FPS    &  \textbf{0.71}/0.29 \\
 \hline
 \textbf{Average} & \textbf{0.74}/0.26 \\
 \hline
\end{tabular}
\caption{User studies results of different methods. We average user preference scores (higher is better). Best results in bold.}

\label{tab:user}
\end{table}

\subsection{Mesh Simplification} 
Meshes are a common way to represent a 3D surface in many
different areas of computer graphics. Similar to point clouds, they are usually composed of thousand points leading to large storage and computational requirements. As described in Section \ref{sec:related}, mesh simplification is a long studied problem that has been tackled only by using greedy algorithms. In this section, we will attempt to propose an alternative method that circumvents the greedy nature of simplification using the simplification technique proposed in Section \ref{sec:method}.

Due to the fact that the proposed model is trained using salient features estimated from the points without taking into account the topology of the initial mesh, we can easily extend the method for the task of mesh simplification. In particular, without the need of any modification to the method, a triangulation algorithm can be used to transform the simplified vertices back to a mesh structure.
The process can be unfolded in two steps. 

\begin{figure}[H]
\begin{center}
   \includegraphics[width=0.99\linewidth]{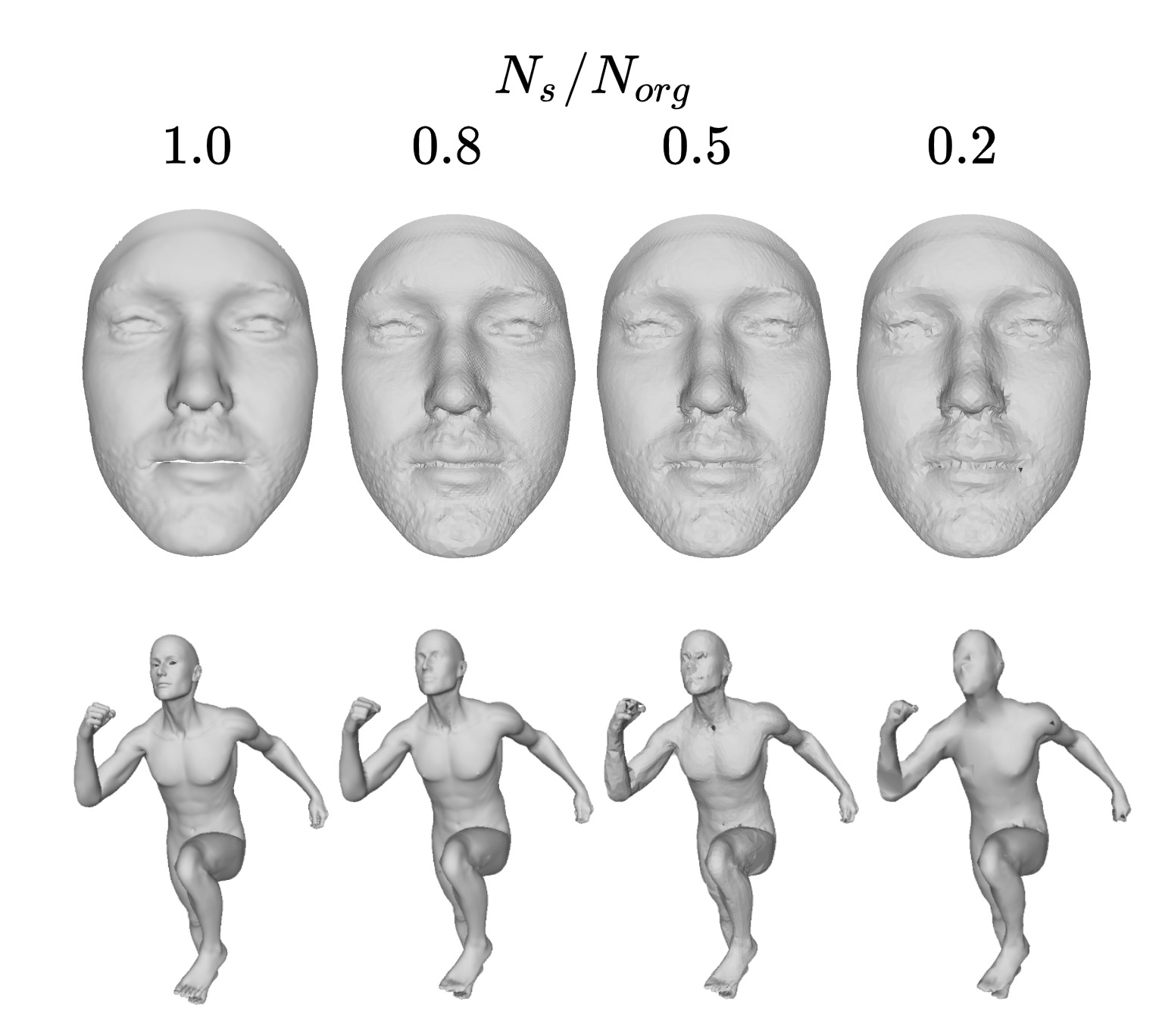}
\end{center}
   \caption{Simplified meshes using the proposed method followed by Ball Pivoting Algorithm.}
\label{fig:mesh}
\end{figure}

Initially, mesh vertices are simplified by treating them as a point cloud but, instead of using a k-nn, the mesh adjacency matrix is utilized in order to determine point connectivity.
In a second step, the remaining vertices are re-triangulated using an off-the-self triangulation algorithm such as Ball Pivoting \citep{bernardini1999ball}. Different triangulation algorithms, such as Delaunay, alpha shapes or Voronoi diagrams, could be also used but we observed that Ball Pivoting algorithm produces better results in small point clouds. Figure \ref{fig:mesh} shows the extension and application of the proposed simplification method to triangular meshes (for various simplification ratios).

\subsection{Computational Time}
\begin{figure}[!ht]
\begin{center}
  \includegraphics[width=\linewidth]{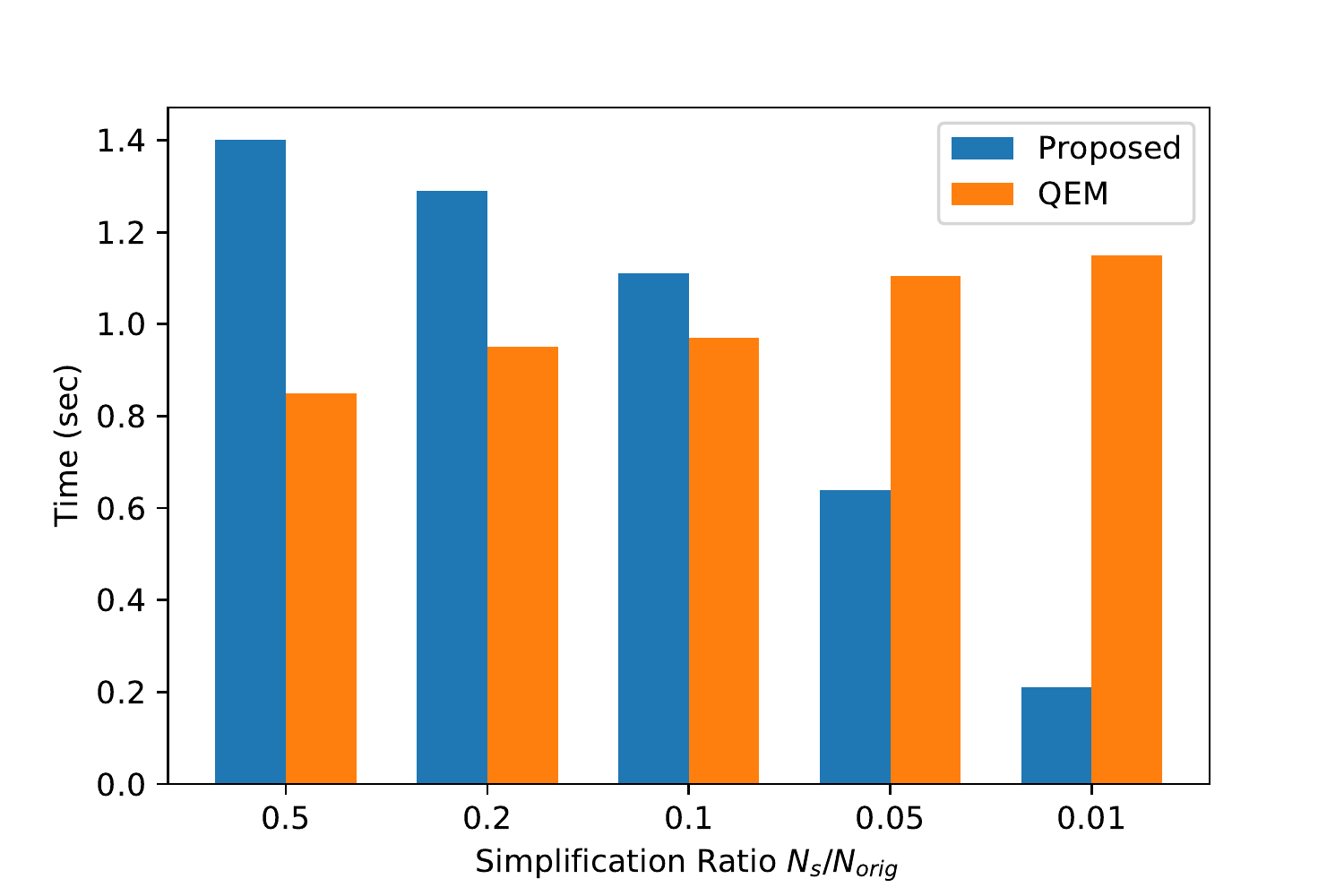}
\end{center}
  \caption{Average time of simplification for the proposed and the QEM methods.}
\label{fig:time}
\end{figure}

Inevitably, in addition to salient point preservation, a proper point cloud simplification method is required to be executed in real-time. Although time complexity is beyond the scope of this study, we assessed the time required for simplifying 80 high-resolution meshes from the TOSCA dataset. 	Since FPS, Uniform and TCP baseline methods do not require any significant computations, we compare the proposed method with the popular QEM approach using a highly optimized version from the MeshLab framework \citep{cignoni2008meshlab}. It is important to note that the code of the proposed method could be further optimized, using parallel programming. Fig. \ref{fig:time} shows that the required mean runtime of the proposed method decreases drastically across all experiments, as the desired simplification increases, requiring just a few seconds to simplify the input point clouds to 1\% of their original size. 

\subsection{Ablation Study: The importance of Adaptive Chamfer Distance}
\begin{table*}[!ht]
\centering
\resizebox{\linewidth}{!}{
\begin{tabular}{l|cccc|cccc|cccc|}

                  & \multicolumn{4}{c}{$N_s/N_{org}=$ 0.2 }                                                                                                & \multicolumn{4}{c}{$N_s/N_{org}=$0.1 }                                                                                                  & \multicolumn{4}{c|}{$N_s/N_{org}=$0.05 }                                                                                                \\
Method            & CD                              & CE                               & RE($\times 10^{-4}$)            & SDM($\times 10^{-4}$)          & CD                              & CE                                & RE($\times 10^{-4}$)            & SDM($\times 10^{-4}$)          & CD                              & CE                               & RE($\times 10^{-4}$)            & SDM($\times 10^{-4}$)           \\ 
\hline

Proposed-CD   & \textbf{1.12}   & 0.40 & 3.96 & 5.52 & \textbf{2.41}  & 0.47 & 4.43  & 9.64  & \textbf{4.91} & 0.58  & 4.99 & 18.3                         \\
Proposed-ACD & 1.15          & 0.39  & 4.01  & 5.51 & 2.54  & 0.46  & 4.42  & 9.61  & 4.97 & 0.56  & 4.96 & 17.9  \\
Proposed-Full   & 1.12  & \textbf{0.37}   & \textbf{3.91}   & \textbf{5.01}                   & 2.45  & \textbf{0.46}                   & \textbf{4.41}                   & \textbf{7.84}                   & 4.93                            & \textbf{{0.57}} & \textbf{4.93}                   & \textbf{16.5}                    \\
\hline
\end{tabular}}
\caption{Ablation study on loss function. Proposed-CD denotes the model trained with CD, Proposed-ACD denotes model trained with adaptive CD and Proposed-Full denotes the model trained with the loss functions introduced in Section 3.3.}
\label{tab:ablation}
\end{table*}
\begin{table*}[!htb]
\centering
\resizebox{\linewidth}{!}{
\begin{tabular}{l|cccccccccccc}

                  & \multicolumn{4}{c}{$N_s/N_{org}=$ 0.2}                                                                                                                                          & \multicolumn{4}{c}{$N_s/N_{org}=$ 0.1}                                                                                                                                          & \multicolumn{4}{c|}{$N_s/N_{org}=$ 0.05}                                                                                                             \\
Method            & CD                             & NC                                    & RE($\times 10^{-4}$)                          & \multicolumn{1}{c|}{SDM($\times 10^{-4}$)}                           & CD                             & NC                                    & RE($\times 10^{-4}$)                          & \multicolumn{1}{c|}{SDM($\times 10^{-4}$)}                           & CD & NC                                    & RE($\times 10^{-4}$)                          & \multicolumn{1}{c|}{SDM($\times 10^{-4}$)}                           \\ \hline
Random            & 2.71                            & 0.37                       & 6.56                                 & \multicolumn{1}{c|}{9.20}                                 & 4.43                                & 0.38                        & 6.74                                 & \multicolumn{1}{c|}{14.39}                                 & 7.78   & 0.39                        & 6.85                             & \multicolumn{1}{c|}{23.63}                                 \\
TCP               & 24.7                               & 0.48                                 & 6.30                               & \multicolumn{1}{c|}{9.27}                                 & 37.2                              & 0.49                               & 6.58                                 & \multicolumn{1}{c|}{14.83}                                 & 53.5  & 0.51                                 & \textcolor{red}{\textbf{6.77 }}                            & \multicolumn{1}{c|}{23.45}                                 \\
FPS               & 2.74                                 & 0.34                    &\textcolor{red}{\textbf{6.25}} & \multicolumn{1}{c|}{9.89}                                 & 4.28                                 & \textcolor{red}{\textbf{0.36}}                                 &\textcolor{red}{\textbf{6.34}} & \multicolumn{1}{c|}{15.86}                                 & 6.83     & 0.39                               &6.81 & \multicolumn{1}{c|}{25.03}                                 \\
QEM               & \textbf{1.92}                                &\textbf{ 0.31}                                & 6.61                        & \multicolumn{1}{c|}{\textcolor{red}{\textbf{9.14}}}                                 & \textbf{2.57}                                 & 0.36                              & 6.81                        & \multicolumn{1}{c|}{\textcolor{red}{\textbf{14.53}}}                                 & \textbf{3.85}    &  \textcolor{red}{\textbf{0.37}} & 6.93                                 & \multicolumn{1}{c|}{\textcolor{red}{\textbf{23.12}}}                                 \\ \hline
Proposed    & \textcolor{red}{\textbf{2.50}}                  & \textcolor{red}{\textbf{0.33}} & {\textbf{6.13}}                         & \multicolumn{1}{c|}{{\textbf{{8.81}}}}                        & \textcolor{red}{\textbf{3.96}}                        &\textbf{0.35} & \textbf{6.24}                               & \multicolumn{1}{c|}{\textbf{14.42}}                        & \textcolor{red}{\textbf{6.46}}     & \textbf{0.37}                                 & \textbf{6.34}                              & \multicolumn{1}{c|}{\textbf{22.31}}                        \\ \hline \hline 
Proposed w/o noise   & 1.12  & 0.29              &3.91                   & \multicolumn{1}{c|}{5.01}               & 2.21  & 0.31                  & 4.41                   & \multicolumn{1}{c|}{7.84}                   & 4.93                            & 0.33 & 4.93                   & \multicolumn{1}{c|}{16.54}                    \\ \hline 
\end{tabular}}
\caption{ Simplification performance tested on TOSCA dataset with addition of Gaussian noise. Best approaches highlighted are highlighted in \textbf{bold} and second best in \textcolor{red}{\textbf{red}}.  "Proposed w/o noise" is reported for reference.}
\label{tab:tosca_noise}
\end{table*}

\begin{figure*}[!ht]
\begin{center}
  \includegraphics[width=\linewidth]{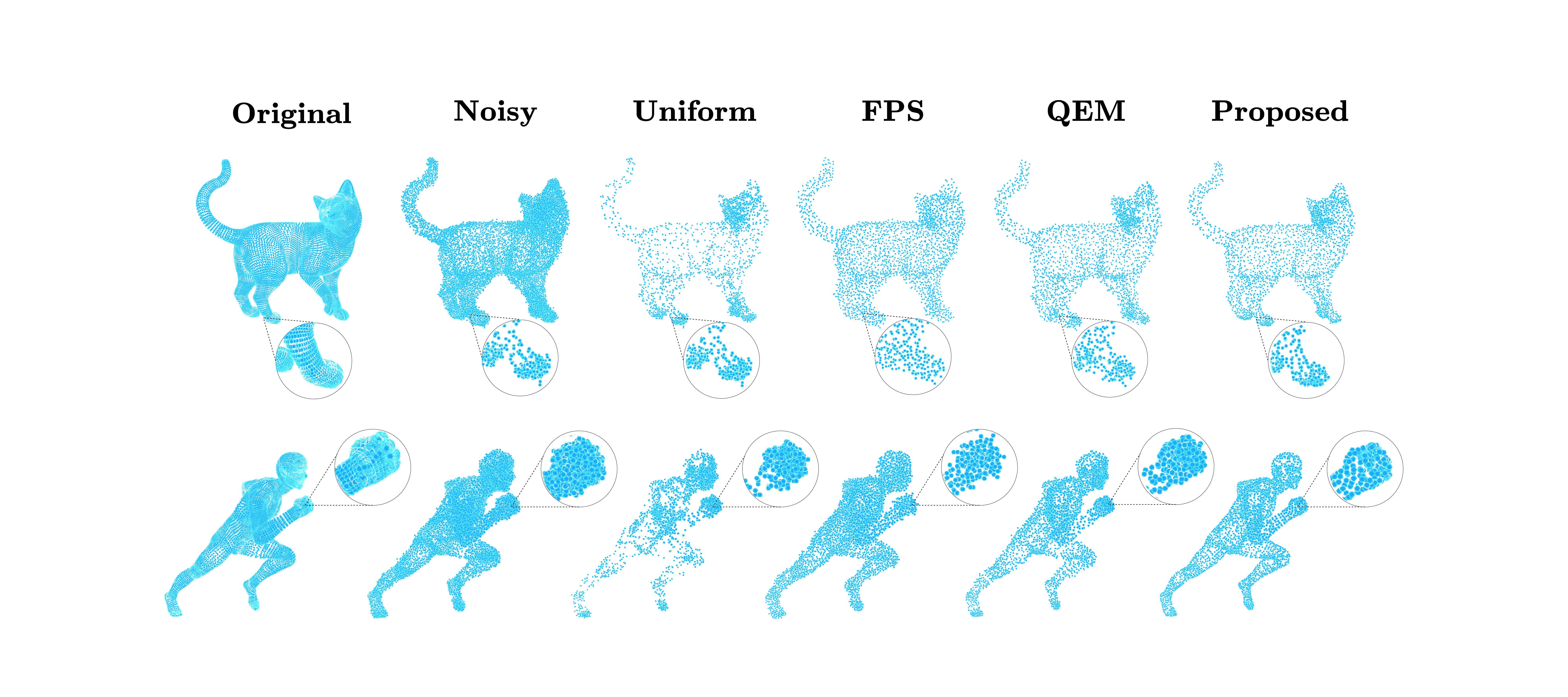}
\end{center}
  \caption{Qualitative comparison between the baseline and the proposed methods for point clouds with Gaussian noise addition. }
\label{fig:noise}
\end{figure*}

As mentioned in Section \ref{losses}, an important component of the proposed simplification framework is the engineering of the curvature guided loss function. In particular, Chamfer Distance (CD) assigns an equal importance weight to each point set, neglecting important points of the point cloud. Thus, semantically meaningful points will be assigned with the same penalty as with points at flat smooth areas. In such way, CD will drive the model to generate smooth results that minimize shape reconstruction without taking into account critical identity details of the object. To break this uniformity, we modified the first term of the CD to assign a different weight to each point according to its curvature. In Table \ref{tab:ablation}, we report the performance of the proposed method trained only with regular CD (Proposed-CD), with adaptive CD (Proposed-ACD), and with both adaptive CD and curvature preservation loss (Proposed-Full). Results reveal that the modified CD exhibits lower perceptual error (CE, RE, SDM) compared to simple CD, while adding a curvature preserving loss (Proposed-Full) further boosts the performance of the model.

\subsection{Simplification under noise conditions}
Although the task of point cloud simplification usually comes after denoising and processing of the raw point clouds, we further examined the behavior of the proposed method under noise conditions and real world point clouds. 

\subsubsection{Gaussian Noise}

To both quantitatively and qualitatively evaluate the performance of the proposed method in the presence of noise we fed the pretrained model with noisy point clouds. In particular, using the TOSCA test set for evaluation, we we added Gaussian noise with unit standard deviation to each point of the point clouds present on the test set. The noisy point clouds along with the reference ones can be seen in the first two columns of Figure \ref{fig:noise}. We fed the noisy point clouds to the proposed model trained on the original TOSCA train set, without further training or tuning. The last three columns of Figure \ref{fig:noise} show and contrast the simplified noisy point clouds generated by the proposed and the baseline methods. Details in sharp areas show that the propose method preserves most of the structure of the input without being affected from the outlier noisy points as much as the baseline methods. In contrast, sampling points directly from the xyz-space using the FPS method produces noisy outputs following the noisy patterns of the inputs. Similarly, QEM selects noisy points  in order to minimize the quadric error of the input planes, such as the outliers in cat's foot and human hand (shown in zoomed areas).

\subsubsection{Real-World Point Clouds}
\begin{figure}[!h]
\begin{center}
  \includegraphics[width=\linewidth]{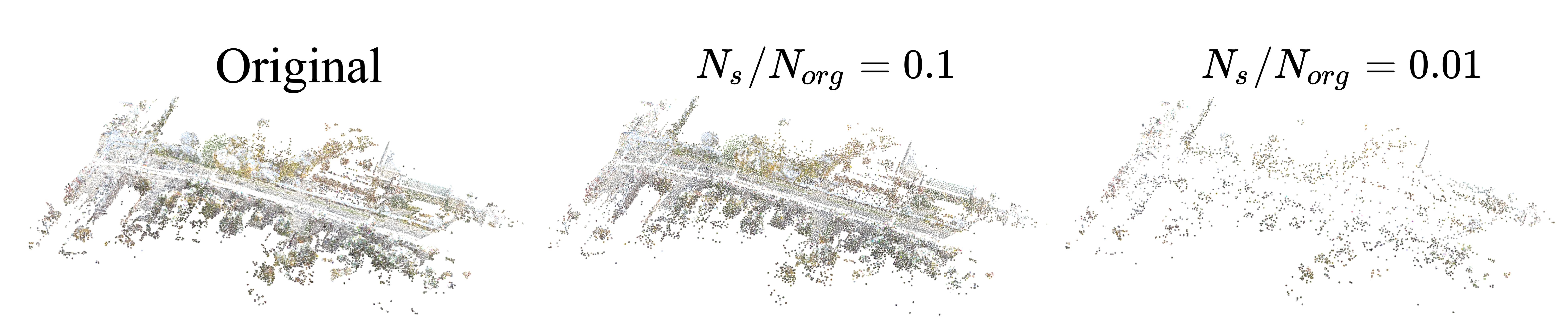}
\end{center}
  \caption{Simplification of real-world scans using the proposed method. Figure better viewed in zoom.}
\label{fig:lidar}
\end{figure}

\begin{table*}[!ht]

\resizebox{\linewidth}{!}{
\centering
\begin{tabular}{l|ccccccccccccccc}
                  \multicolumn{15}{c}{Torronto3D}                                                                           \\
                  & \multicolumn{5}{c}{$N_s/N_{org}=$ 0.2}                                                                                                                                          & \multicolumn{5}{c}{$N_s/N_{org}=$ 0.1}                                                                                                                                          & \multicolumn{5}{c|}{$N_s/N_{org}=$ 0.05}                                                                                                             \\
Method            & CD     & NC   & CE($\times 10^{-2}$)  & RE($\times 10^{-4}$)  & \multicolumn{1}{c|}{SDM($\times 10^{-4}$)}  & CD   & NC  & CE($\times 10^{-2}$)  & RE($\times 10^{-4}$) & \multicolumn{1}{c|}{SDM($\times 10^{-4}$)}                           & CD & NC   & CE($\times 10^{-2}$)  & RE($\times 10^{-4}$)  & \multicolumn{1}{c|}{SDM($\times 10^{-4}$)}                           \\ \hline

Uniform          &  0.31 &  0.577 & 8.15 & 11.27 & \multicolumn{1}{c|}{0.47} &
		  0.62 &  0.634 & 8.89 & 11.56 & \multicolumn{1}{c|}{0.48} &
 		  1.27 &  0.679 & 9.15 & 11.91 &\multicolumn{1}{c|}{0.48}
   \\

TCP    &        5.90 &  0.894 & 15.64 & 14.47 & \multicolumn{1}{c|}{0.58} &
		7.40 &  0.912 & 12.41 & 12.95 & \multicolumn{1}{c|}{0.53}  &
		9.58 &  0.912 & 13.02 & 12.61 & \multicolumn{1}{c|}{0.53}
  \\

FPS     &   \textbf{0.17} &  \textbf{0.509} & 6.42 & 11.22 & \multicolumn{1}{c|}{0.46} &
	 \textbf{0.34} &  \textbf{0.565}    & 7.01 & 11.32 & \multicolumn{1}{c|}{0.46} & 
	\textbf{0.70} &  \textbf{0.619}    & 7.51 & 11.37 & \multicolumn{1}{c|}{0.47} 
                   \\

 \hline

Proposed    &   0.18 &   0.512  & \textbf{5.67} & \textbf{11.02} & \multicolumn{1}{c|}{\textbf{0.34}} &
	       0.37 &   0.595  & \textbf{6.35} & \textbf{11.10} & \multicolumn{1}{c|}{\textbf{0.38}} & 
	     0.75 &   0.644  &  \textbf{6.88} & \textbf{11.15} & \multicolumn{1}{c|}{\textbf{0.41}} 
                   \\  \hline \hline

\end{tabular}}
\caption{ Simplification performance tested on outdoor point cloud from Torronto3D dataset. Best approaches highlighted are highlighted in \textbf{bold}. The proposed method model is trained with TOSCA dataset.}
\label{tab:lidar}
\end{table*}
A significant application of point cloud simplification methods is to sub-sample points of real-world scanners that generate million of points from the representative surface. To test the performance of the proposed method on such scenario, we utilized Torronto3D dataset \citep{tan2020toronto3d}, containing outdoor point clouds acquired with LIDAR sensors. Again, we utilized the pretrained model on TOSCA dataset, without further training or tuning. Quantitative results summarized in Table \ref{tab:lidar} demonstrate that the proposed method outperforms baseline methods in perceptual quality measures (CE, RE, SDM). Although FPS method achieved the lower CD and NC errors, it produces smooth results that minimize the overall shape loss without preserving essential details of the object. Figure \ref{fig:lidar} shows examples of the simplified lidar point clouds at different simplification ratios generated by the proposed method. 

\section{Conclusion}
Our work emphasises on the proposal of a learnable, neural-based simplification technique to substitute and overcome the inefficiencies of traditional greedy simplification methods. In this study we presented the first, to the best of our knowledge, learnable point cloud simplification method that aims at preserving salient features while at the same time retaining the global structural appearance of the input 3D object. In order to ensure that both salient and shape features of the input are preserved, we modified chamfer distance to penalize high curvature points. The proposed method is composed by three modules: the projection network, the point selector and the refinement layer. The projection network embeds the points of the input point cloud to a higher dimensional space that is sampled from the point selector module. Finally, the refinement layer, slightly modifies the position of the selected points to minimize the curvature error.  The proposed method can simplify a point cloud to 1\% of its original size in real-time, addressing the literature limitations regarding computational complexity. 

As shown in an extensive series of both quantitative and qualitative experiments the proposed method outperforms its counterparts under most perceptual criteria. We have also explored the zero-shot capabilities of the model and assessed its performance on out-of-distribution shapes or even to noisy and real-world point clouds. Results proved that the proposed model can be used off-the-self without the need of further training or tuning. 
Regarding future work, we plan to adapt the proposed method to mesh structures using a more sophisticated triangulation process. In particular, instead of using off-the-shelf triangulation algorithms on top of the point cloud simplification model, we aim to extend the proposed method to predict the triangulation of the simplified model utilizing the priors of the input mesh. 

\bibliography{sn-article}


\end{document}